\documentclass[letterpaper, 10 pt, conference]{ieeeconf}
\IEEEoverridecommandlockouts
\usepackage[utf8]{inputenc}
\usepackage[T1]{fontenc}
\usepackage{hyperref}
\usepackage{balance}
\usepackage{graphicx}
\usepackage[font=small,labelfont=bf]{caption} 
\usepackage{subcaption}
\usepackage{xspace}
\usepackage{url}

\usepackage{adjustbox}
\usepackage{amsfonts} 
\usepackage{array} 
\usepackage{colortbl} 
\usepackage{xcolor}
\usepackage{algorithm}
\usepackage{algpseudocode}
\algdef{SE}[DOWHILE]{Do}{doWhile}{\algorithmicdo}[1]{\algorithmicwhile\ #1}%
\algnewcommand\algorithmicinput{\textbf{Input:}}
\algnewcommand\algorithmicoutput{\textbf{Output:}}
\algnewcommand\Input{\item[\algorithmicinput]}
\algnewcommand\Output{\item[\algorithmicoutput]}
\algnewcommand{\Continue}{\textbf{continue}}
\algnewcommand{\Break}{\textbf{break}}
\usepackage{tcolorbox}

\newcommand{\Npos}{\mathbb{N}_{>0}}

\newcommand{\frameworkName}{PhyRoGen\xspace}
\title{\fontsize{23}{28}\selectfont \frameworkName: Synthetic Generation of Physical Manipulation Puzzles Using Procedural Content Generation}
\title{\fontsize{20}{24}\selectfont \frameworkName: Synthetic Generation of Physical Robot Manipulation Puzzles Using Procedural Content Generation}
\author{Lennart Julian Droß$^{1}$ and Andreas Orthey$^{1}$ and Marc Toussaint$^{1,2}$
\thanks{This work has been supported by German Federal Ministry of Research, Technology, and Space (BMFTR) under the Robotics Institute Germany (RIG).}%
\thanks{$^{1}$ Technical University of Berlin, Germany}%
\thanks{$^{2}$ Robotics Institute Germany}%
}

\begin{document}

\maketitle
\begin{abstract}
    
Robot manipulation of physical puzzles is important for automatic assembly and disassembly tasks.
However, to enable robots to solve physical puzzles, manipulation skills need to be learned, which requires large training datasets, the generation of which is often time consuming and tedious.
To overcome this problem, we propose the Physical Robot Manipulation Puzzle Generation framework (PhyRoGen), which leverages procedural content generation (PCG) for automated generation of synthetic datasets of manipulation puzzles.
PhyRoGen is a general-purpose puzzle generator, which can generate physical puzzles with interlocking object dependencies, where one articulated object must be manipulated before another can be moved.
Based upon PhyRoGen, we define six concrete generators which we use to generate 24 physical puzzles. 
By using a benchmarking framework, we are able to solve all puzzles in 1 to 300 seconds using sampling-based planning algorithms.
Finally, we demonstrate that every generated puzzle is manipulatable by using a KUKA LBR iiwa robot in a physical simulation.
This shows that our framework is able to procedurally generate unique, solvable robot manipulation puzzles, which is a crucial ingredient to benchmark manipulation algorithms and to develop robust foundation models.

\end{abstract}
\section{Introduction}


Large datasets are essential for developing and evaluating robot manipulation algorithms, both for benchmarking motion planners~\cite{Chamzas2022RAL, moll2015benchmarking} and for training learning-based systems~\cite{kroemer2021review, firoozi2025foundation}.
However, manually curating such datasets is tedious, limits the diversity of generated scenarios, and makes it difficult to produce systematic variations~\cite[p.~2]{gaiSurvey}.

To tackle this problem, we leverage procedural content generation (PCG) to algorithmically create synthetic datasets.
While some works have proposed PCG to generate datasets for robot manipulation problems~\cite{deitke2022procthor, Chamzas2022RAL}, they do not generate problems with \emph{interlocking object dependencies}.
These are problems where one articulated object must be manipulated before another can be moved. For example, a door (with a revolute joint) may need to be opened before a cube (on a prismatic joint) behind it can be moved.
Throughout this paper, the term \emph{joint} refers to the kinematic connection between object links in the puzzle (e.g., a door hinge or a sliding rail), not to the joints of the robot manipulator.
We call such problems \emph{physical manipulation puzzles}~\cite{baum2017opening,Bayraktar2023RAL}.

\begin{figure}
    \centering
    \includegraphics[width=0.48\linewidth, height=0.48\linewidth]{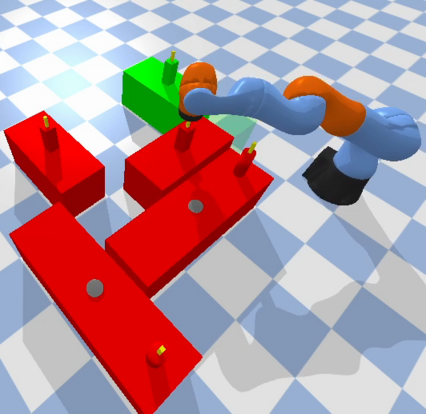}
    \includegraphics[width=0.48\linewidth, height=0.48\linewidth]{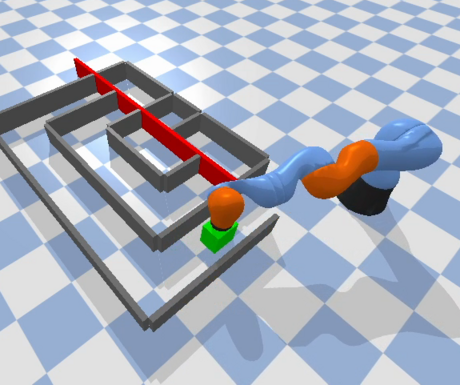}
    \caption{A KUKA mobile manipulator solves a procedurally generated manipulation puzzle with interlocking object dependencies.
    \textbf{Left:} The grid world environment, where a green slider has to be moved, but is blocked by multiple red objects.
    \textbf{Right:} The Move N Times environment, where a green cube has to be moved out of a maze which is blocked by a sliding prismatic door, requiring multiple re-grasping sequences.}
    \label{fig:pullfigure}
\end{figure}

To create manipulation puzzles, we introduce the \emph{Phy}sical \emph{Ro}bot Manipulation Puzzle \emph{Gen}eration Framework (\frameworkName).
\frameworkName consists
of a general maker algorithm, which takes as input sequences of objects, joints, transformations, and returns as output a kinematic chain representing a manipulation puzzle. By using this framework, we show that a variety of puzzles can be created, including lockbox puzzles~\cite{baum2017opening}, simple escape rooms~\cite{Bayraktar2023RAL}, and puzzles which require re-grasping sequences~\cite{simeon2004manipulation}. In total, we create $24$ puzzles, and showcase that they are solvable using a benchmark framework to compare planning algorithms~\cite{moll2015benchmarking, Chamzas2022RAL}. Finally, we devise a method to manipulate those puzzles using a KUKA LBR iiwa robot. This is shown in Fig.~\ref{fig:pullfigure}, where we showcase two puzzles generated with \frameworkName. The mobile manipulator robot is able to manipulate the objects to move the green object to a desired goal configuration, while removing red objects whenever they block the execution.

To summarize, our contributions are the following:
\begin{itemize}
    \item Puzzle generation: We develop the \frameworkName puzzle generation framework and implement it using Blender~\cite{blender} in combination with the Phobos~\cite{phobos2020} plugin.
    \item Puzzle benchmarking: We create six generators to produce a total of $24$ puzzles, and benchmark them using sampling-based puzzle-solving algorithms using Robowflex~\cite{kingston2022robowflex, lee2018dart} and OMPL~\cite{sucan2012open}.
    \item Puzzle manipulation: We show how puzzles can be manipulated and solved using a simulated KUKA LBR iiwa mobile robot platform using PyBullet~\cite{coumans2021pybullet} and OMPL~\cite{sucan2012open}.
\end{itemize}

\begin{figure*}
    \centering
    \includegraphics[width=\linewidth]{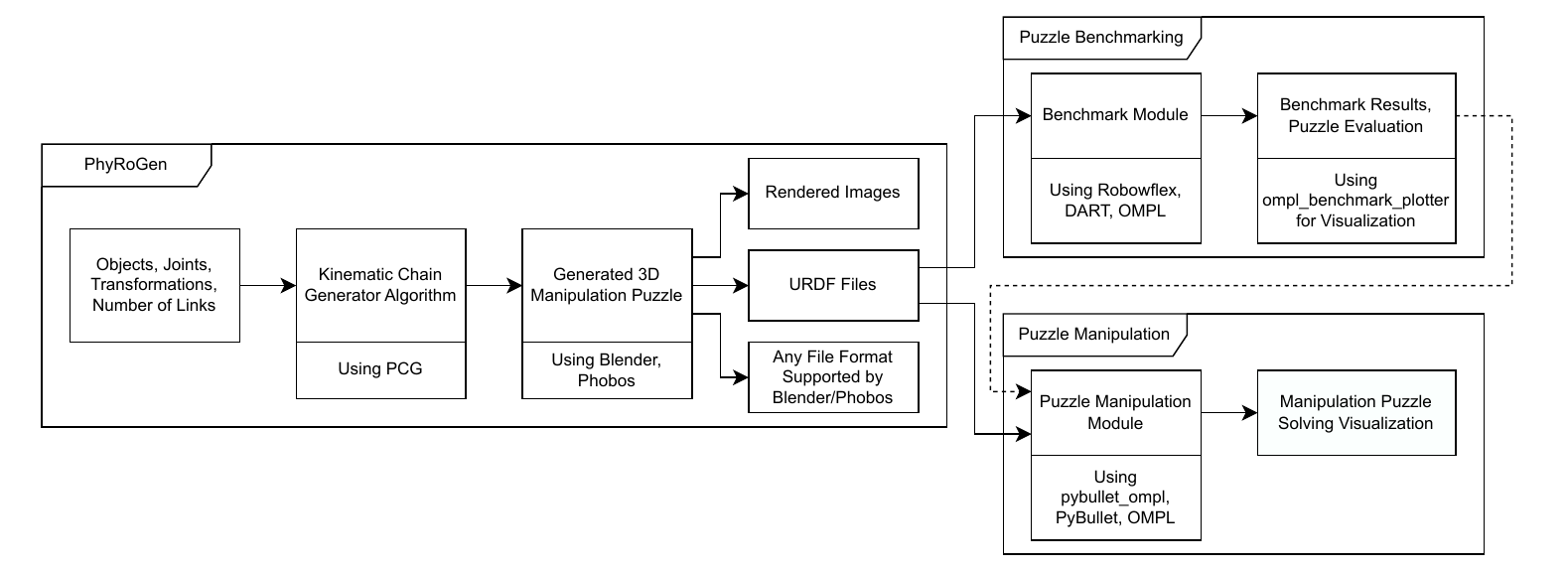}
    \caption{Systems overview about this paper, including the puzzle generation (left), and the puzzle benchmarking and manipulation modules (right). }
    \label{fig:system}
\end{figure*}

A complete overview of the modules involved is shown in Fig.~\ref{fig:system}. All material related to this work, including puzzle URDF files, images, benchmark results (graphs and database files), manipulation videos, and source code is freely available\footnote{Overview page: \href{https://sites.google.com/view/robot-manipulation-puzzles}{sites.google.com/view/robot-manipulation-puzzles}}.

\section{Related Work}

PCG~\cite{togelius2011procedural} has been used in the creation of digital games since the 1980s with games like \textit{Rogue} which use procedurally generated environments~\cite{pcgbook}. This has lead to the genre of \emph{roguelike} games~\cite{harris2020exploring}, which include the creation of virtual game spaces~\cite{viana2021procedural}, game bits (textures, sound, game objects), game systems (ecosystems, road networks, entity behavior), and game scenarios (puzzles, levels, stories)\cite{hendrikx2013procedural, pcgbook}.
Recently, PCG systems have been used to generate complete game universes like in \textit{No Man's Sky} or \textit{Minecraft}~\cite{herve2021comparing, awiszus2022wor}.

To avoid repetitions and keep players engaged, game development has also focused on procedural puzzle generation~\cite{dekegel2019procedural}. 
An exemplary physical puzzle is \textit{Sokoban}~\cite{zakaria2022procedural}, a PSPACE-complete problem~\cite{culberson1997sokoban}, which has caught the attention of the gaming community~\cite{kartal2016data, taylor2015procedural, karman2018generating, zakaria2022procedural}. 
While Sokoban puzzles present interesting challenges for robots, they are not well suited for physical robot exploration, because unrecoverable states can be easily reached and are difficult to recognize. 

Games also play an important role as learning environments for reinforcement learning (RL) agents~\cite{Sutton2018reinforcement}.
This includes Arcade games~\cite{bellemare2013arcade, mnih2015human} and real-time strategy games~\cite{vinyals2019grandmaster}.
To generalize those RL systems, it becomes important to generate synthetic worlds.
Systems like Procgen~\cite{cobbe2020leveraging} or MiniHack~\cite{samvelyan2021minihack} can generate different game environments with simple point-based reward functions.
Despite their usefulness for RL agents, those systems are not suitable for robotics due to the focus on mostly 2-dimensional spaces and simplified action spaces.

In the last decade, robot learning has also become a central focus of the robotics community~\cite{kroemer2021review, xiao2025robot}. Similarly to RL agents, robot learning requires large datasets to create robust systems~\cite{kober2013reinforcement}. 
To generate such datasets, researchers have focused on perturbating existing problems with automatic domain randomization~\cite{openai2019rubiks}. 
However, to generalize to unseen environments, it becomes crucial to be able to learn on automatically generated worlds. Therefore, several generator algorithms have been proposed like ProcTHOR~\cite{deitke2022procthor}, which is a PCG framework for interactive physical environments, MotionBenchMaker~\cite{Chamzas2022RAL}, a tool to generate motion planning datasets, or ClutterGen~\cite{jia2024cluttergen}, which can generate diverse sets of tables with objects for manipulation. 
Those datasets can be used in learning benchmarks like Meta-World~\cite{yu2020meta} or RLBench~\cite{james2020rlbench}, which are specifically designed to be used in robotics frameworks.
Those generators, however, do not generate environments with interlocking object dependencies, where one articulated object blocks another.
This makes them less suitable for task and motion planning (TAMP) that involves physical exploration.

While robot learning algorithms have mostly been applied to robot manipulation tasks~\cite{openai2019rubiks, kroemer2021review}, work on learning for TAMP algorithms~\cite{kaelbling2011hierarchical, garrett2021integrated} has only recently been started~\cite{pmlr-v202-driess23a}. 
In TAMP, an algorithm has to reason over both discrete, logical constraints and continuous state spaces connecting them~\cite{toussaint2018differentiable, kingston2019exploring}. 
One category of TAMP algorithms tackles physical puzzles~\cite{baum2017opening}, where objects have dependencies preventing them from being manipulated in a random order~\cite{Bayraktar2023RAL}.
This makes physical puzzles differ from tabletop scenarios~\cite{gao2022fast} or top-down rearrangement planning scenarios~\cite{tsesmelis2024re}.

Our proposed puzzle generator is complementary to both learning and TAMP planners in that \frameworkName creates environments with interlocking object dependencies, such that manipulation puzzles can be created, where a robot has to reason about object interactions~\cite{Bayraktar2023RAL}.
This is an important requirement both for learning frameworks and for advancing TAMP algorithms.
\section{\frameworkName: Physical Robot Manipulation Puzzle Generation Framework}

The \frameworkName framework consists of three parts. First, we define kinematic chain generators, which are functions to generate a sequence of kinematic chains from an integer seed. Based upon them, we detail the \frameworkName maker algorithm, which is a general-purpose kinematic chain generator maker, taking as input objects, joints, and transformations, and returning a kinematic chain. Finally, we depict six concrete kinematic chain generators, which we use to build $24$ scenarios of randomly generated manipulation puzzles.

\subsection{Kinematic Chain Generators}

A kinematic chain generator is a function which gets as input an integer \emph{seed} in
$\mathbb{N}$, plus a set of objects, a set of joints, a set of transformations, and a number of objects. As output, a generator produces a kinematic chain consisting of links and joints together with their relative transformations. We define a generator to have the following attributes:

\begin{itemize}
  \item \textbf{Deterministic}. For a given seed and identical input, the generator always produces the same kinematic chain.
  \item \textbf{Interdependent}. Each generator produces kinematic chains which represent puzzles which are solvable only if all joints are actuated. That means there are no superfluous joints added.
  \item \textbf{Solvable and Manipulatable}. Each kinematic chain should lead to a puzzle generated which is solvable and eventually manipulatable by a mobile manipulator.
\end{itemize}

\subsection{\frameworkName Maker Algorithm}

Here we present the \frameworkName maker algorithm. This is an iteration-based algorithm, which aims to create a new object in each iteration, and which produces as output a kinematic chain representing a physical puzzle.

%

\newcommand{\var}[1]{\texttt{#1}}
\newcommand{\joint}{\var{jnt}}
\newcommand{\object}{\var{obj}}
\newcommand{\transform}{\var{tf}}
\newcommand{\kinematicchain}{\var{K}}
\newcommand{\kinematicchaintmp}{\var{Kt}}

\newcommand{\Objects}{\var{Objects}}
\newcommand{\seed}{\var{Seed}}
\newcommand{\Items}{\var{Items}}
\newcommand{\Item}{\var{Item}}
\newcommand{\Joints}{\var{Joints}}
\newcommand{\Transforms}{\var{Transforms}}
\newcommand{\IndComment}[1]{\item[] \hspace*{-0.2cm} {// \textit{#1}}}

\begin{algorithm}[t]
\caption{\frameworkName Maker}\label{Algorithm1}
\begin{algorithmic}[1]
  \Input{$(\Objects,\Joints,\Transforms, N, \seed)$}
  \Output{Kinematic Chain $\kinematicchain$}

  \State $\kinematicchain \gets \Call{InitializeKinematicChain}{ }$
  \For{$n \in \{1,\ldots,N\}$} 
    \While {True}
    \State $\object \gets \Call{GetItem}{\Objects, n, \seed}$
    \State $\joint \gets \Call{GetItem}{\Joints, n, \seed}$
    \State $\transform \gets \Call{GetItem}{\Transforms, n, \seed}$
    \State $\kinematicchaintmp \gets \Call{NewKinematicChain}{\kinematicchain, \object, \joint, \transform}$
    \IndComment{Joint should invalidate puzzle in its default state}
    \State $\kinematicchaintmp \gets \Call{FixJoint}{\kinematicchaintmp, \joint}$
    \If {\Call{ValidatePuzzle}{$\kinematicchaintmp$}}
        \State \Continue
    \EndIf
    \IndComment{Puzzle should be valid when joint is variable}
    \State $\kinematicchaintmp \gets \Call{UnfixJoint}{\kinematicchaintmp, \joint}$
    \If {\textbf{Not} \Call{ValidatePuzzle}{\kinematicchaintmp}}
        \State \Continue
    \EndIf
    \State $\kinematicchain \gets \kinematicchaintmp$
    \State \Break
    \EndWhile
  \EndFor
  \State \Return \kinematicchain
\end{algorithmic}
\end{algorithm}

The complete algorithm is detailed in Algorithm~\ref{Algorithm1}. As input, we are given the tuple ($\Objects$, $\Joints$, $\Transforms$, $N$, $\seed$), whereby $\Objects$ is either a
finite set of objects or an infinite sequence of objects, $\Joints$ is either a finite
set of joints or an infinite sequence of joints, $\Transforms$ is either a finite set
of transformations or an infinite sequence of transformations, $N$ is the number of
objects of the algorithm, and $\seed$ is the unique seed in $\Npos$ which acts as the randomization, i.e. which objects, joints, and transformations to use in each iteration. As output, the algorithm returns a kinematic chain $\kinematicchain$, representing a physical puzzle. 

To achieve this, the following steps are carried out. 
First, we initialize an empty kinematic chain (Line 1). 
We then run $N$ iterations of the algorithm, whereby in each iteration we enter a while loop (Line 3) which ends when a new kinematic chain has been found (Line 16--17). 
To find a new kinematic chain, we first get a random object $\object$, joint $\joint$, and transformations $\transform$ from the input sequences (Line 4--6). 
The \texttt{GetItem} function either choose a random element from a set by using the seed, or returns the $n$-th element in the sequence. 
Those items are used to extend the current kinematic chain $\kinematicchain$ to create a new temporary kinematic chain $\kinematicchaintmp$ (Line 7). 
We then conduct two validations steps. 
First, we check if the puzzle is solvable when the new object is at its default position (Line 8--11). 
If yes, we reject this solution and continue, since the object would not contribute to a solution and would thereby be superfluous. 
This validation step uses the RRT-Connect planner~\cite{kuffner2000rrt} on the temporary kinematic chain $\kinematicchaintmp$ until a first solution is found or a timeout is reached.
Second, we check if there exists a solution to the puzzle when the object is movable along its joint (Line 12--15). 
If there exists no solution, then the puzzle is not solvable, and we have to reject the kinematic chain. 
Once the new kinematic chain has passed both checks, we update the global kinematic chain $\kinematicchain$ (Line 16) and exit the while loop (Line 17). 
After all iterations have passed, we return the kinematic chain (Line 20). 

For every chain generated, we define a physical puzzle problem as the problem of moving the first
link from its default position to a dedicated goal position. There are
three cases: If the object has an attached prismatic joint, we define the goal
as the opposite end of its limits. If a revolute joint is present, we define the
goal as a $90$ degree rotation from its current position in either clockwise or
counterclockwise direction. If the joint is $SE(2)$, we define the goal as being
a random placement outside of a bounding box around the final generated puzzle.

\subsection{Six Kinematic Chain Generators}

Given the general-purpose the \frameworkName maker algorithm, we present six concrete input tuples, each leading to a different set of kinematic chain generators.
The chosen six input tuples of objects, joints, transformations, and number of objects
are shown in Table~\ref{table:generator-sequences}. We have indicated in this table infinite sequences as gray colored cells and finite sets as white colored cells. This input directly defines the kinematic chain generators through \frameworkName, and can be used to generate an infinite sequence of manipulation puzzles by changing the input seed.

The kinematic chains are built directly in the 3D modeling software Blender~\cite{blender} using a Python script~\cite{conlan2017blender}. From this 3D description, we can both render images of the puzzles and export them to the Unified Robot Description Format (URDF) by using the Blender extension Phobos~\cite{phobos2020}.  

\newcolumntype{L}[1]{>{\raggedright\arraybackslash\hspace{0.5em}}p{#1}}
\newcolumntype{C}[1]{>{\centering\arraybackslash}p{#1}} 
\setlength{\tabcolsep}{0pt} 
\renewcommand{\arraystretch}{1.5} 
\definecolor{highlightcolor}{gray}{0.9}

\begin{table*}[t]
\centering
\caption{Input Sequence Configurations. Gray cells represent infinite sequence inputs, while the remaining white cells represent finite sets.\label{table:generator-sequences}}
\begin{tabular}{|L{0.16\textwidth}|L{0.27\textwidth}|L{0.27\textwidth}|L{0.17\textwidth}|C{0.1\textwidth}|} 
\hline
\textbf{Name} & \textbf{Objects} & \textbf{Joints} & \textbf{Transformations} & \textbf{Number of Objects} \\ \hline
\textbf{Simple Slider} & \{T-shaped slider\} & \{Prismatic\} & $\{0, 180\}$ & Seed+1 \\ \hline
\textbf{Continuous Space} & \{Cuboid\} & \{Revolute\} & $[0,90] \times [0,1]$ & 4 \\ \hline
\textbf{Grid World} & \{Cuboid\} & \{Revolute, Prismatic\} & $\{0, 90, 180, 270\}$ & 5 \\ \hline
\textbf{Lockbox Random} & \cellcolor{highlightcolor}\{Cuboid, Disk, Cuboid, Disk, \ldots\} & \cellcolor{highlightcolor}\{Prismatic, Revolute, Prismatic, \ldots\} & $\{0, 90, 180, 270\}$ & 7 \\ \hline
\textbf{Move N Times} & \cellcolor{highlightcolor}\{Cuboid, Door(N), Corridor(1), \ldots\} & \cellcolor{highlightcolor}\{SE(2), Prismatic, Fixed, Fixed, \ldots\} & \cellcolor{highlightcolor}$\{0, 180, 0, \ldots\}$ & Seed+2 \\ \hline
\textbf{Rooms} & \{Room, Door Room\} & \{Fixed\} & $\{0, 90, 180, 270\}$ & 4 \\ \hline
\end{tabular}
\end{table*}
\newcommand{\SeedShift}{1.3cm}

\begin{figure*}[!htbp]
\centering
\captionsetup[subfigure]{labelformat=simple, labelsep=colon, font=small}

\begin{minipage}[t]{0.02\textwidth}
    \vspace*{-\SeedShift}\rotatebox{90}{\parbox{1cm}{\raggedleft Seed 1}}
\end{minipage}\hfill
\begin{subfigure}[t]{0.15\textwidth}
    \centering
    \includegraphics[width=\linewidth]{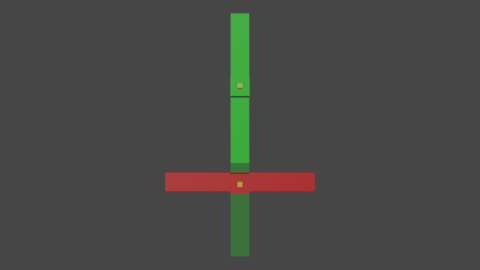}
    \caption*{Simple Slider 1}
\end{subfigure}\hfill
\begin{subfigure}[t]{0.15\textwidth}
    \centering
    \includegraphics[width=\linewidth]{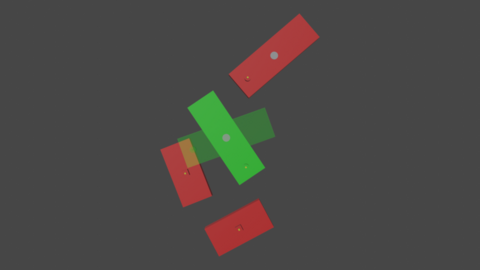}
    \caption*{Continuous Space 1}
\end{subfigure}\hfill
\begin{subfigure}[t]{0.15\textwidth}
    \centering
    \includegraphics[width=\linewidth]{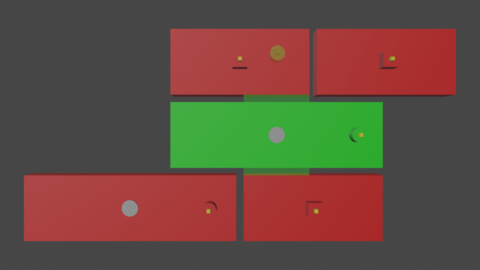}
    \caption*{Grid World 1}
\end{subfigure}\hfill
\begin{subfigure}[t]{0.15\textwidth}
    \centering
    \includegraphics[width=\linewidth]{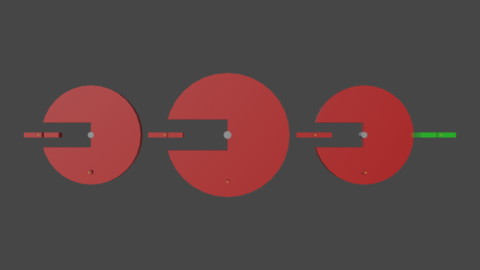}
    \caption*{Lockbox Random 1}
\end{subfigure}\hfill
\begin{subfigure}[t]{0.15\textwidth}
    \centering
    \includegraphics[width=\linewidth]{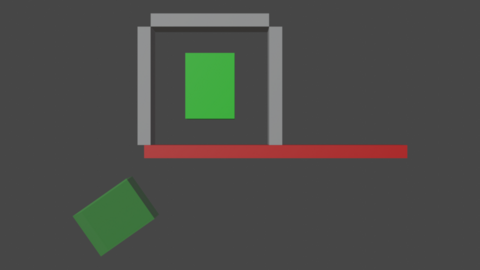}
    \caption*{Move N Times 1}
\end{subfigure}\hfill
\begin{subfigure}[t]{0.15\textwidth}
    \centering
    \includegraphics[width=\linewidth]{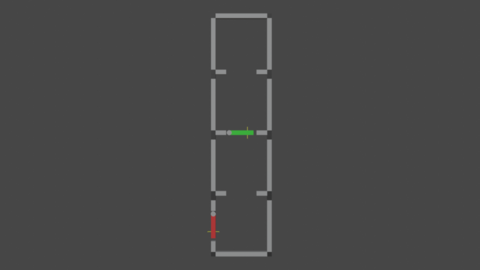}
    \caption*{Rooms 1}
\end{subfigure}

\vspace{2mm}

\begin{minipage}[t]{0.02\textwidth}
    \vspace*{-\SeedShift}\rotatebox{90}{\parbox{1cm}{\raggedleft Seed 2}}
\end{minipage}\hfill
\begin{subfigure}[t]{0.15\textwidth}
    \centering
    \includegraphics[width=\linewidth]{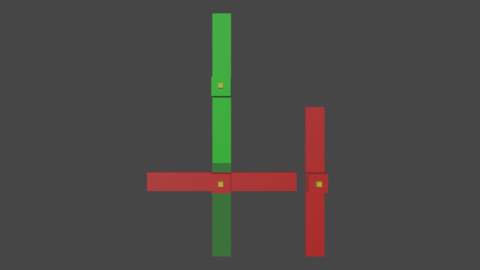}
    \caption*{Simple Slider 2}
\end{subfigure}\hfill
\begin{subfigure}[t]{0.15\textwidth}
    \centering
    \includegraphics[width=\linewidth]{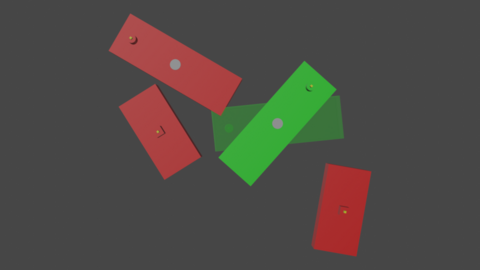}
    \caption*{Continuous Space 2}
\end{subfigure}\hfill
\begin{subfigure}[t]{0.15\textwidth}
    \centering
    \includegraphics[width=\linewidth]{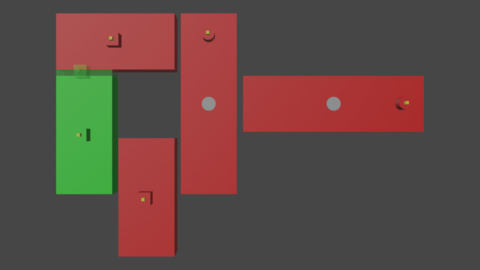}
    \caption*{Grid World 2}
\end{subfigure}\hfill
\begin{subfigure}[t]{0.15\textwidth}
    \centering
    \includegraphics[width=\linewidth]{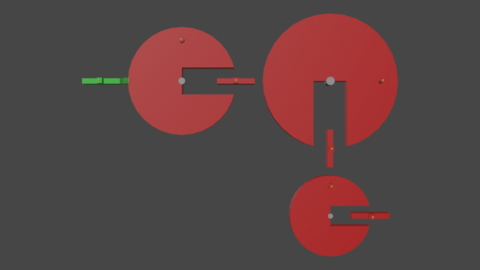}
    \caption*{Lockbox Random 2}
\end{subfigure}\hfill
\begin{subfigure}[t]{0.15\textwidth}
    \centering
    \includegraphics[width=\linewidth]{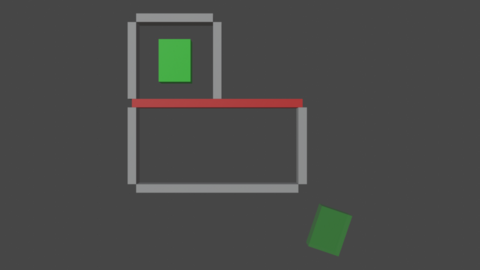}
    \caption*{Move N Times 2}
\end{subfigure}\hfill
\begin{subfigure}[t]{0.15\textwidth}
    \centering
    \includegraphics[width=\linewidth]{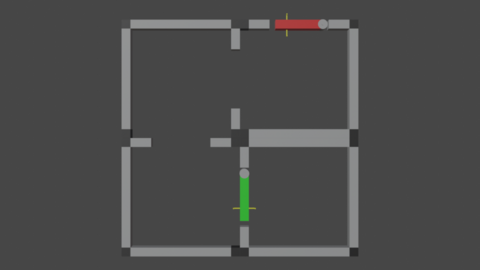}
    \caption*{Rooms 2}
\end{subfigure}

\vspace{2mm}

\begin{minipage}[t]{0.02\textwidth}
    \vspace*{-\SeedShift}\rotatebox{90}{\parbox{1cm}{\raggedleft Seed 3}}
\end{minipage}\hfill
\begin{subfigure}[t]{0.15\textwidth}
    \centering
    \includegraphics[width=\linewidth]{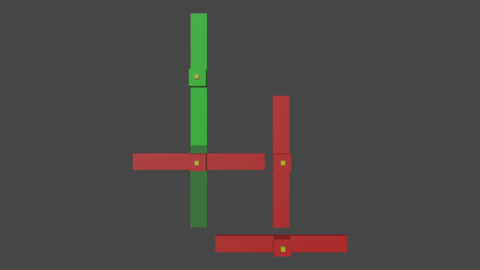}
    \caption*{Simple Slider 3}
\end{subfigure}\hfill
\begin{subfigure}[t]{0.15\textwidth}
    \centering
    \includegraphics[width=\linewidth]{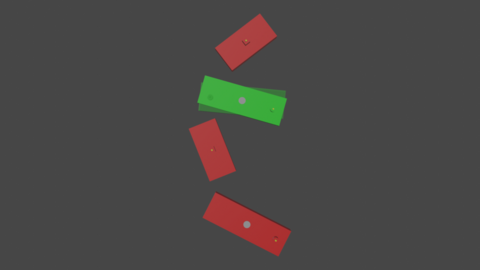}
    \caption*{Continuous Space 3}
\end{subfigure}\hfill
\begin{subfigure}[t]{0.15\textwidth}
    \centering
    \includegraphics[width=\linewidth]{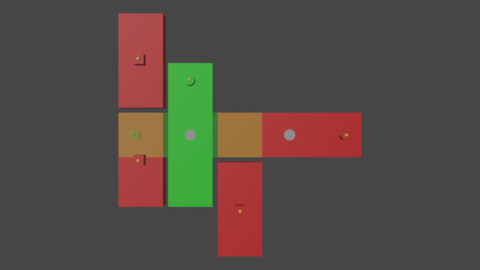}
    \caption*{Grid World 3}
\end{subfigure}\hfill
\begin{subfigure}[t]{0.15\textwidth}
    \centering
    \includegraphics[width=\linewidth]{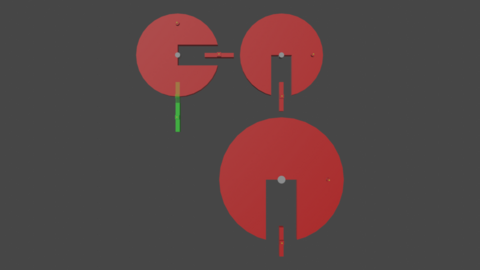}
    \caption*{Lockbox Random 3}
\end{subfigure}\hfill
\begin{subfigure}[t]{0.15\textwidth}
    \centering
    \includegraphics[width=\linewidth]{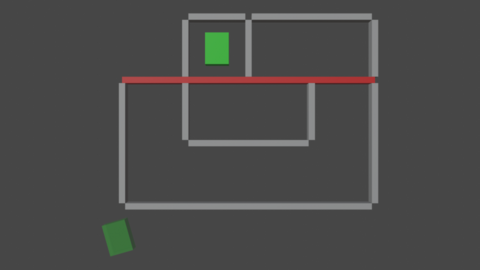}
    \caption*{Move N Times 3}
\end{subfigure}\hfill
\begin{subfigure}[t]{0.15\textwidth}
    \centering
    \includegraphics[width=\linewidth]{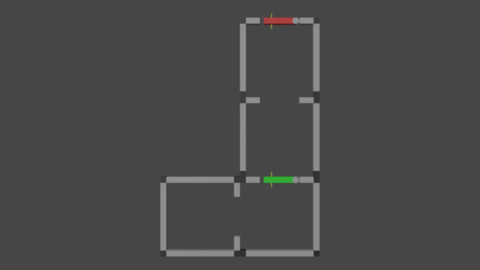}
    \caption*{Rooms 3}
\end{subfigure}

\vspace{2mm}

\begin{minipage}[t]{0.02\textwidth}
    \vspace*{-\SeedShift}\rotatebox{90}{\parbox{1cm}{\raggedleft Seed 4}}
\end{minipage}\hfill
\begin{subfigure}[t]{0.15\textwidth}
    \centering
    \includegraphics[width=\linewidth]{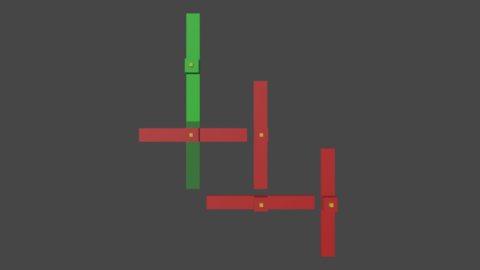}
    \caption*{Simple Slider 4}
\end{subfigure}\hfill
\begin{subfigure}[t]{0.15\textwidth}
    \centering
    \includegraphics[width=\linewidth]{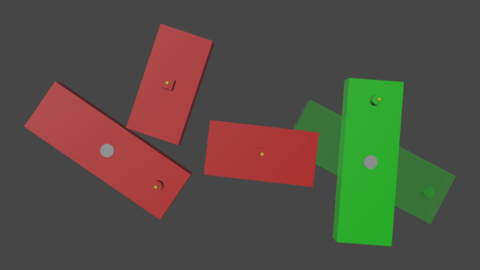}
    \caption*{Continuous Space 4}
\end{subfigure}\hfill
\begin{subfigure}[t]{0.15\textwidth}
    \centering
    \includegraphics[width=\linewidth]{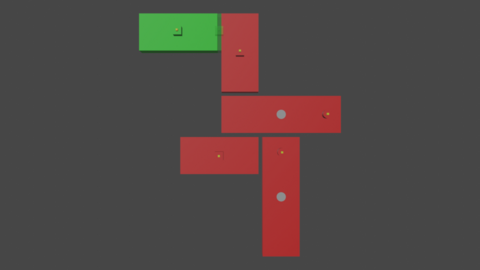}
    \caption*{Grid World 4}
\end{subfigure}\hfill
\begin{subfigure}[t]{0.15\textwidth}
    \centering
    \includegraphics[width=\linewidth]{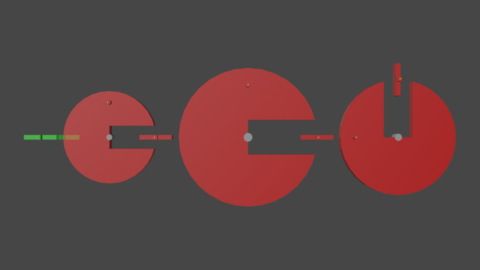}
    \caption*{Lockbox Random 4}
\end{subfigure}\hfill
\begin{subfigure}[t]{0.15\textwidth}
    \centering
    \includegraphics[width=\linewidth]{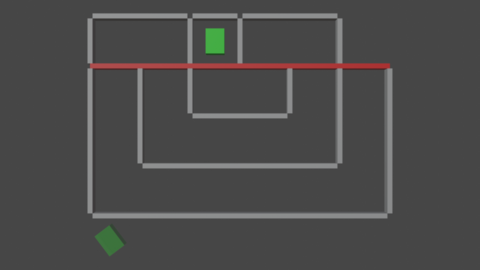}
    \caption*{Move N Times 4}
\end{subfigure}\hfill
\begin{subfigure}[t]{0.15\textwidth}
    \centering
    \includegraphics[width=\linewidth]{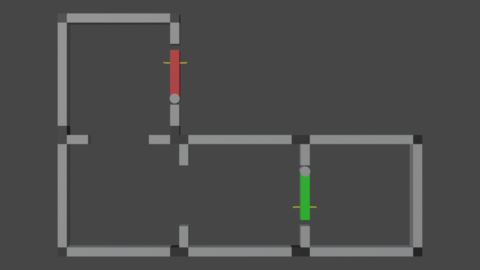}
    \caption*{Rooms 4}
\end{subfigure}

\caption{Visualization of puzzle environments for the first four seeds using the input from Table~\ref{table:generator-sequences}.}
\label{fig:puzzle_environments}
\end{figure*}



\section{Puzzle Benchmarks and Puzzle Manipulation\label{sec:benchmark-and-manip}}

Once all puzzles have been created, we use a puzzle solver to find solutions and execute them using a robot manipulator. To this end, we use sampling-based motion planning methods~\cite{Orthey2023AnnualReview} to solve the puzzles, whereby we rely on the less-action RRT~\cite{Bayraktar2023RAL} (LA-RRT), which is a multi-object planner to compute a minimal number of moves to solve a given puzzle. 

To enable this, we provide an open source implementation. 
This implementation uses Robowflex~\cite{kingston2022robowflex}, and combines it with the Open Motion Planning Library (OMPL)~\cite{sucan2012open}, using the internal benchmarking tools \cite{moll2015benchmarking}.

To show that the generated puzzles are not only solvable, but can also be manipulated, we use a custom method to manipulate them. This method works as follows. First we import the results of the lowest-cost run of LA-RRT for each puzzle (after the time limit) and extract both the object sequences and the object paths. For each object path, we compute an inverse kinematics (IK) solution for the robot at designated grasp points on the objects. To execute those motions, we compute how the grasp point moves when the object is moved along its path. We then use the end-effector Jacobian~\cite{craig2017introduction} to generate a joint direction at the grasp point which moves the object along the computed object path. To deal with singularities, we observe possible joint flips on the robot during the execution. If a singularity occurs, we release the grasp, and re-grasp at a new grasp position at which the robot is not in a singularity at the next object position. This process is continued until the object has attained its final position.

Implementation-wise, we use a KUKA LBR iiwa on a mobile platform for manipulation of the puzzles. We simulate this in PyBullet~\cite{coumans2021pybullet} with an interface to OMPL~\cite{sucan2012open}. 
All computations are using the internal IK solver of PyBullet to compute joint configurations at grasp points. To plan connections to grasp points, we use the Batch Informed Trees (BIT*)~\cite{gammell2020batch} planner in OMPL. BIT* is used for two tasks. First, to reach the IK solutions, which have been found at a grasp point. Second, to compute motions to new grasp points after singularities are detected. This is a slightly different problem since we need a motion from a grasp point to another grasp point.

\begin{figure*}
  \centering
  \begin{subfigure}[b]{0.24\textwidth}
    \centering
    \includegraphics[width=\textwidth]{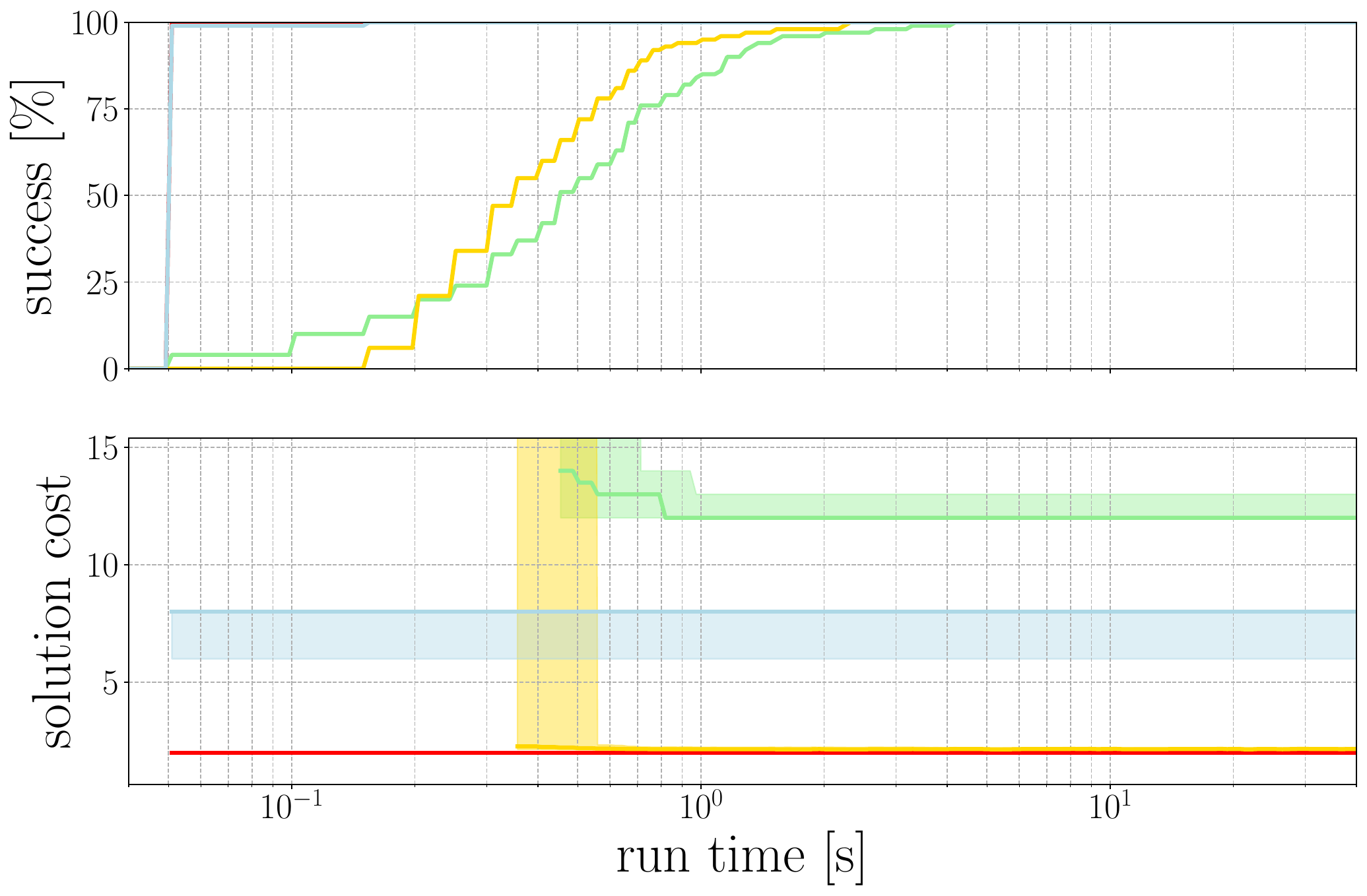}
    \caption*{Simple Sliders 1}
  \end{subfigure}\hfill
  \begin{subfigure}[b]{0.24\textwidth}
    \centering
    \includegraphics[width=\textwidth]{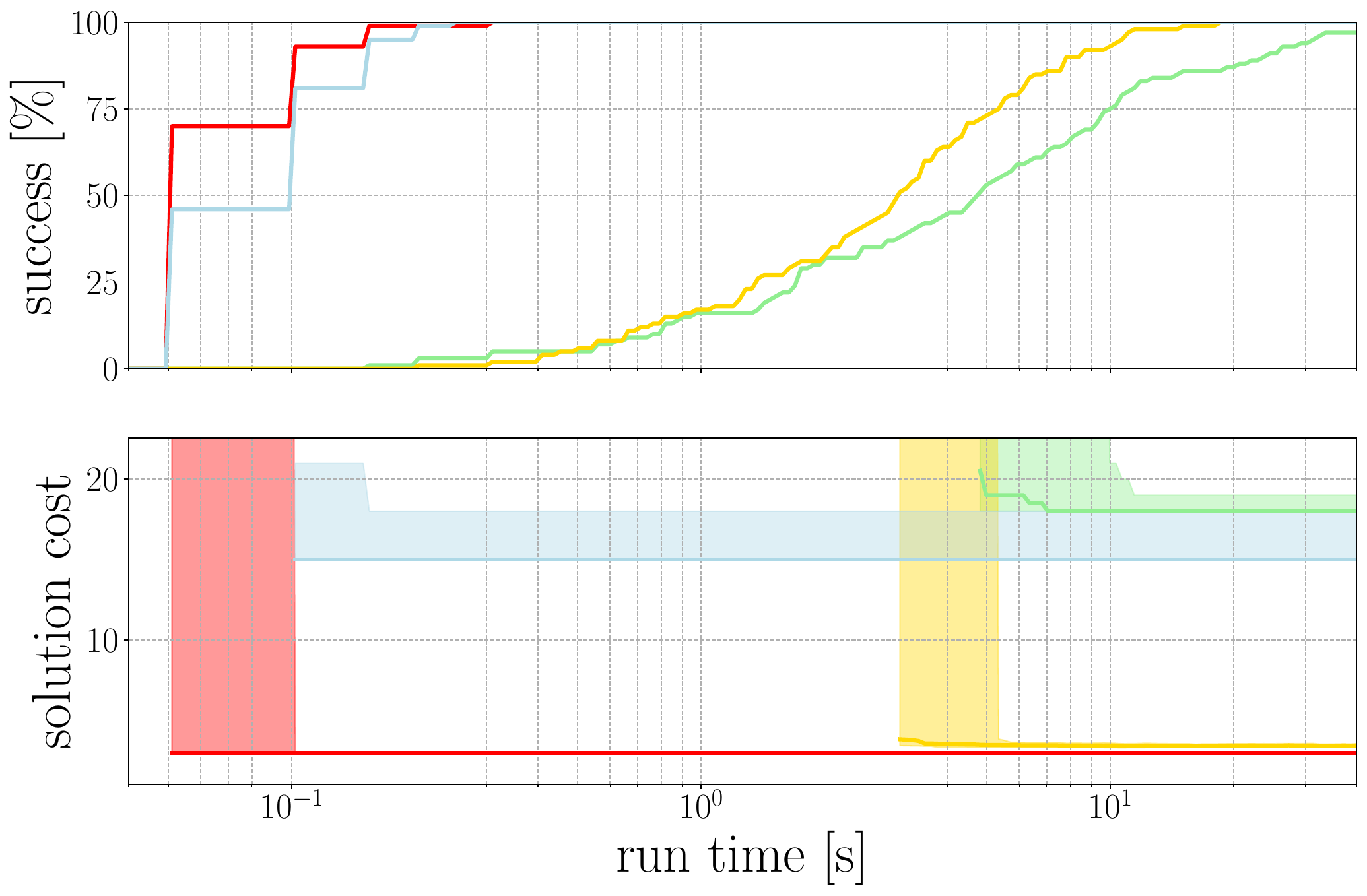}
    \caption*{Simple Sliders 2}
  \end{subfigure}\hfill
  \begin{subfigure}[b]{0.24\textwidth}
    \centering
    \includegraphics[width=\textwidth]{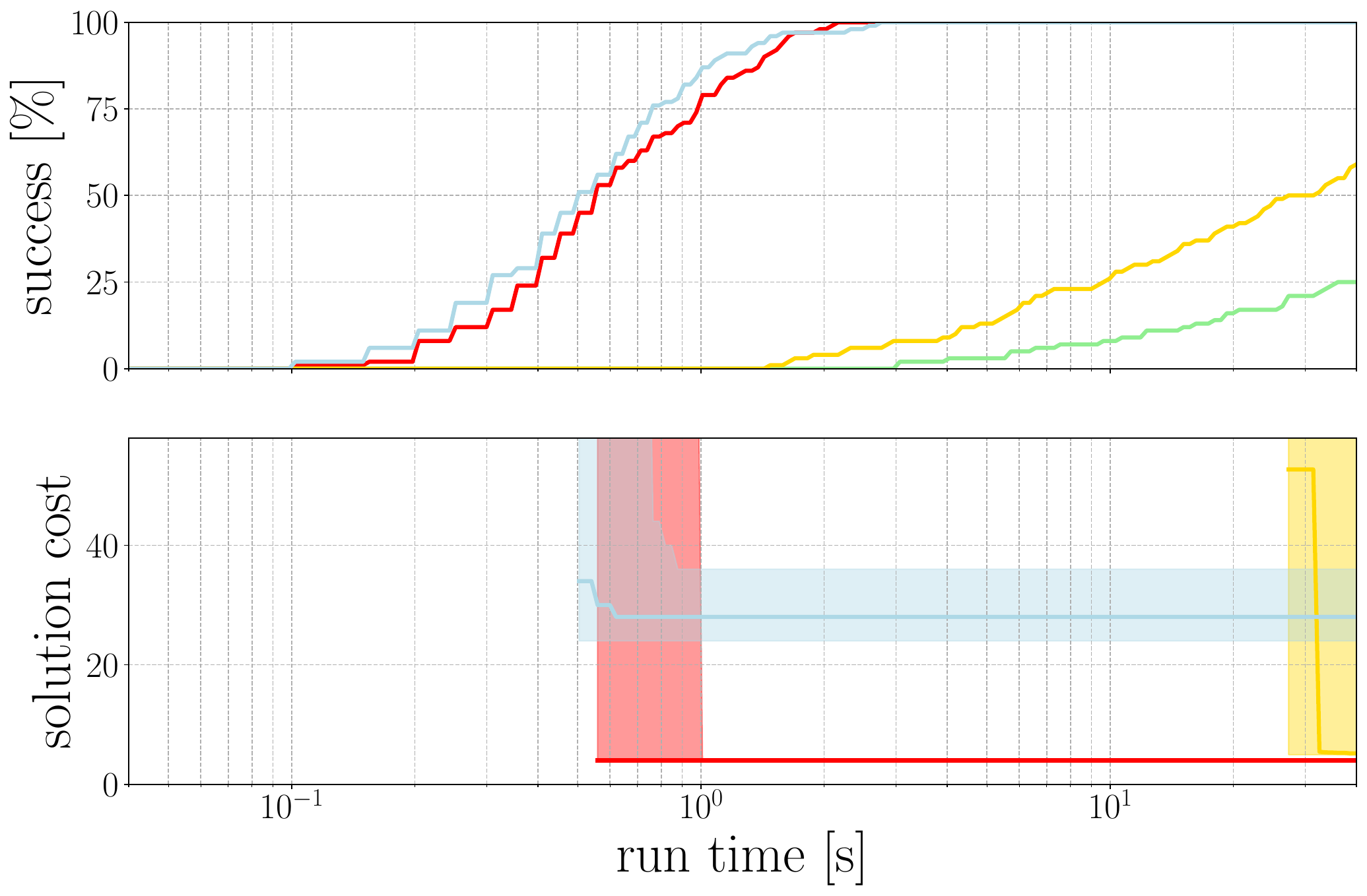}
    \caption*{Simple Sliders 3}
  \end{subfigure}\hfill
  \begin{subfigure}[b]{0.24\textwidth}
    \centering
    \includegraphics[width=\textwidth]{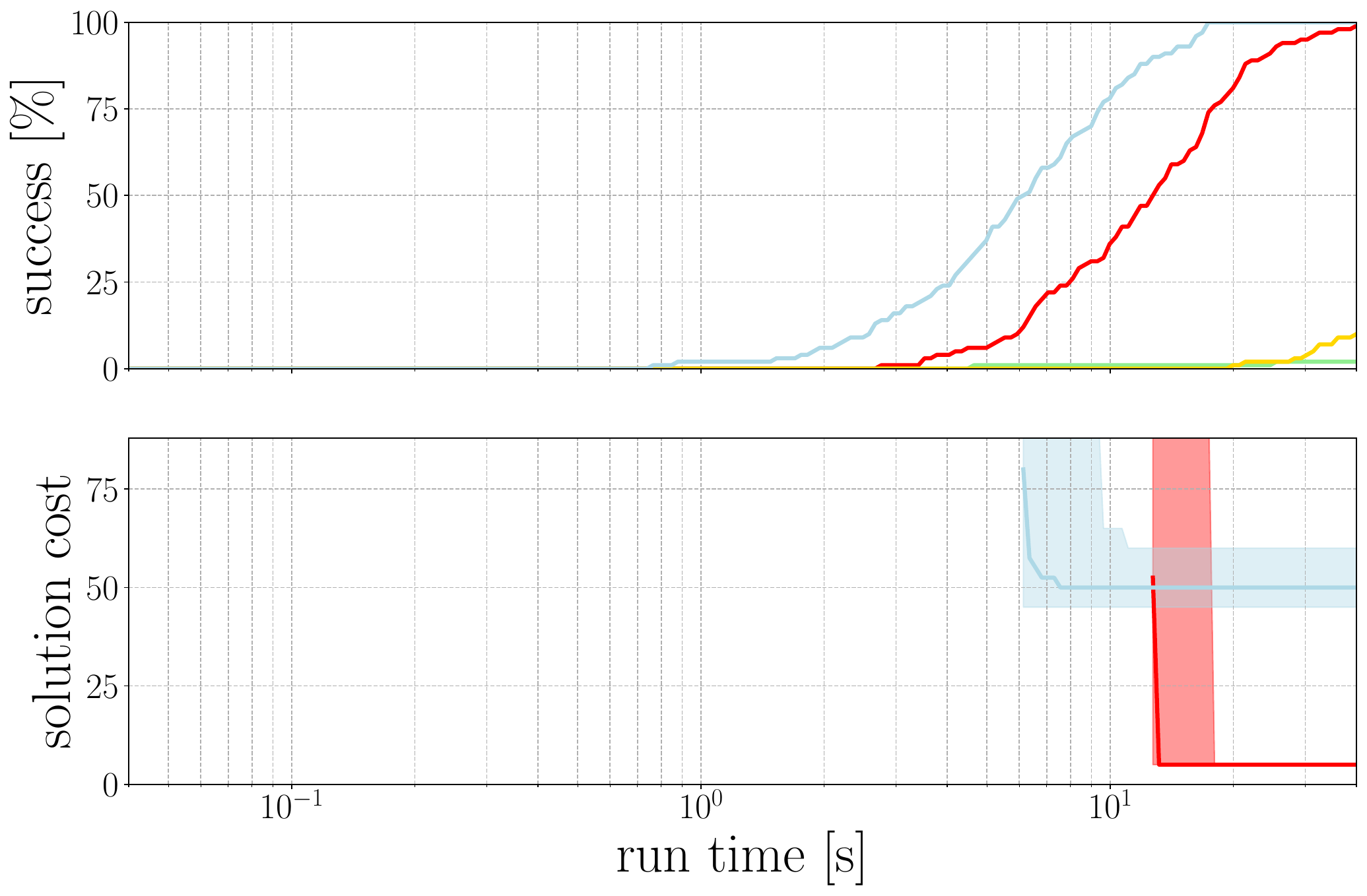}
    \caption*{Simple Sliders 4}
  \end{subfigure}\\[1ex]
  
  \begin{subfigure}[b]{0.24\textwidth}
    \centering
    \includegraphics[width=\textwidth]{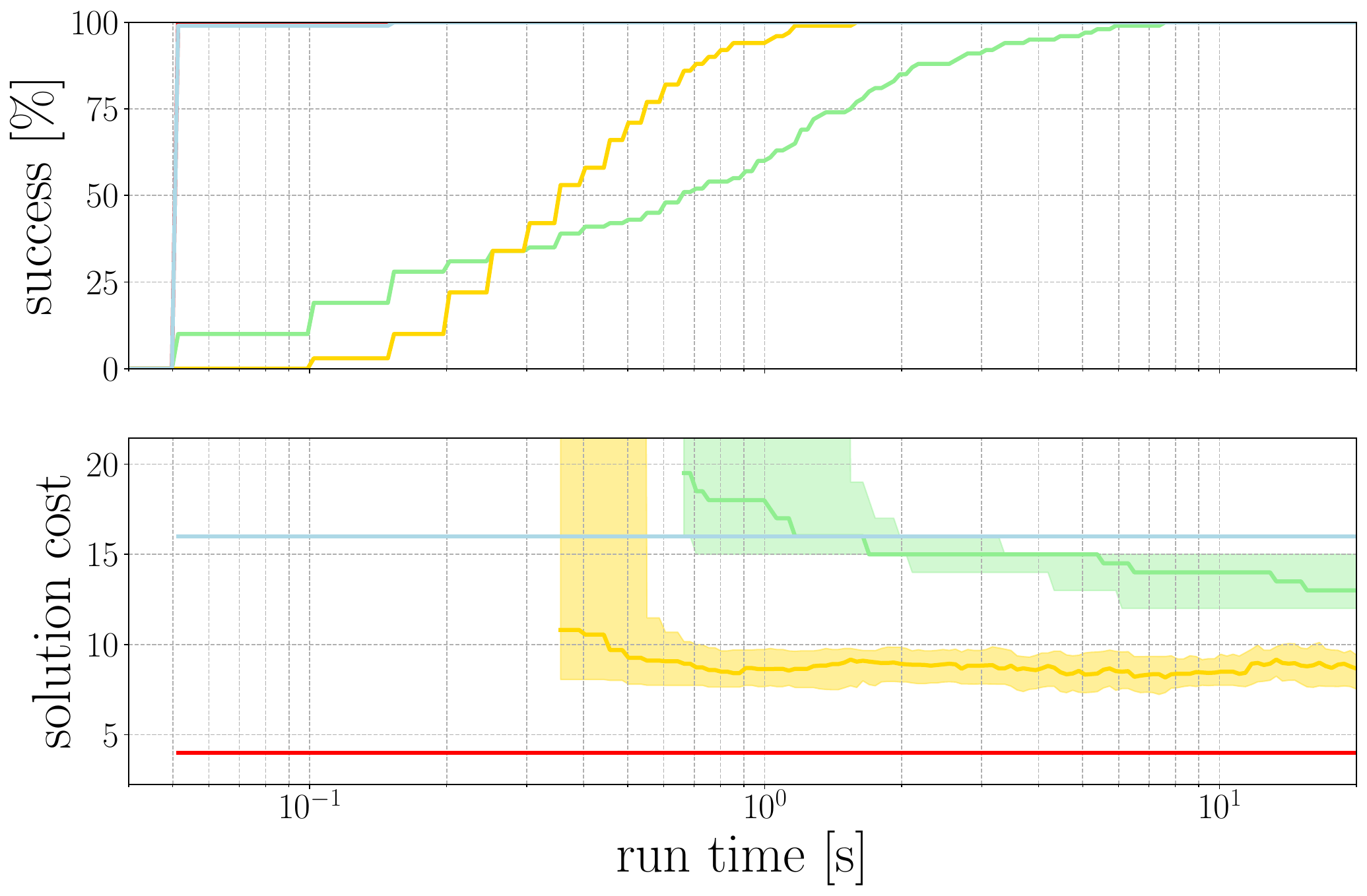}
    \caption*{Continuous Space 1}
  \end{subfigure}\hfill
  \begin{subfigure}[b]{0.24\textwidth}
    \centering
    \includegraphics[width=\textwidth]{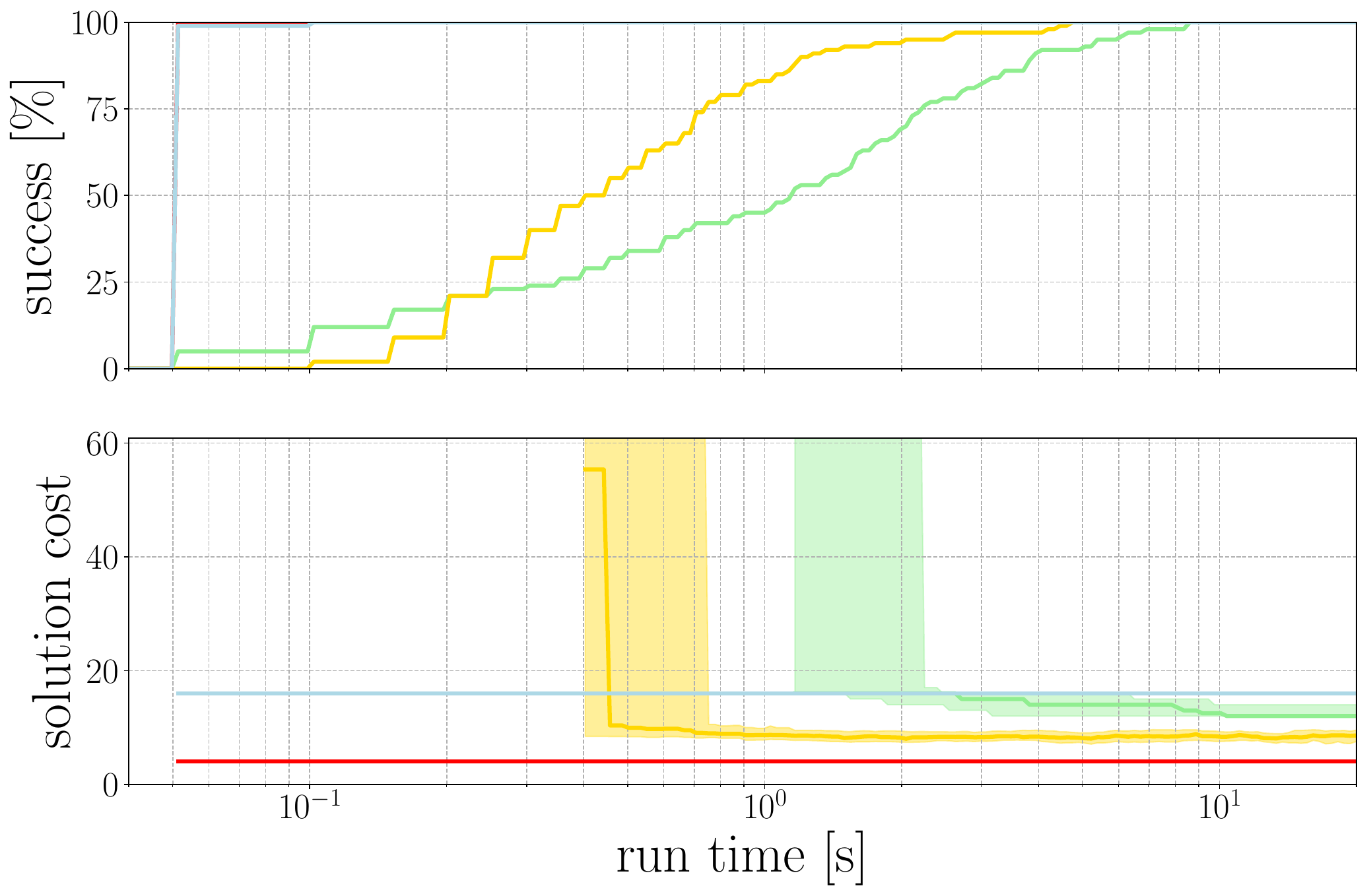}
    \caption*{Continuous Space 2}
  \end{subfigure}\hfill
  \begin{subfigure}[b]{0.24\textwidth}
    \centering
    \includegraphics[width=\textwidth]{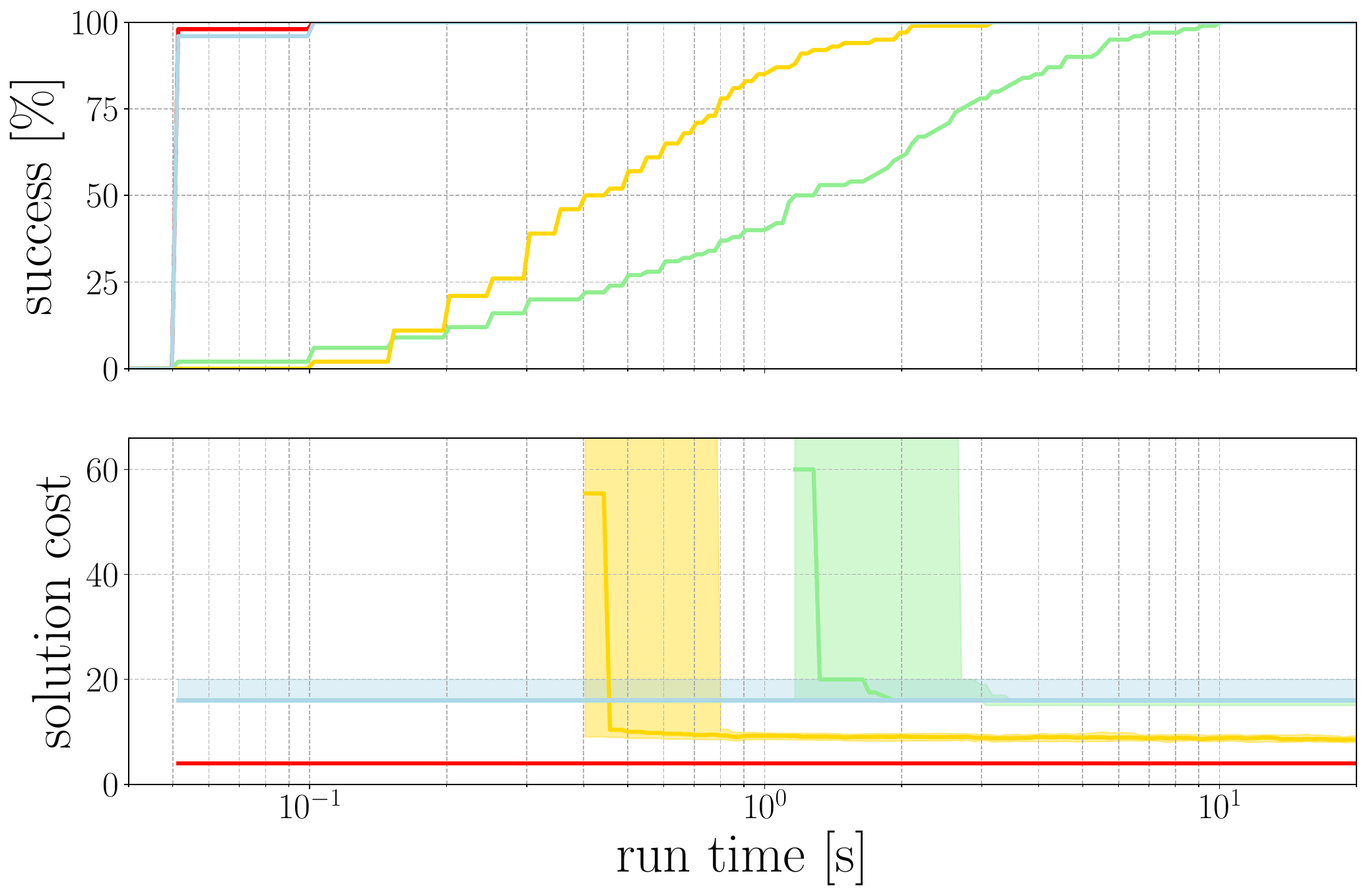}
    \caption*{Continuous Space 3}
  \end{subfigure}\hfill
  \begin{subfigure}[b]{0.24\textwidth}
    \centering
    \includegraphics[width=\textwidth]{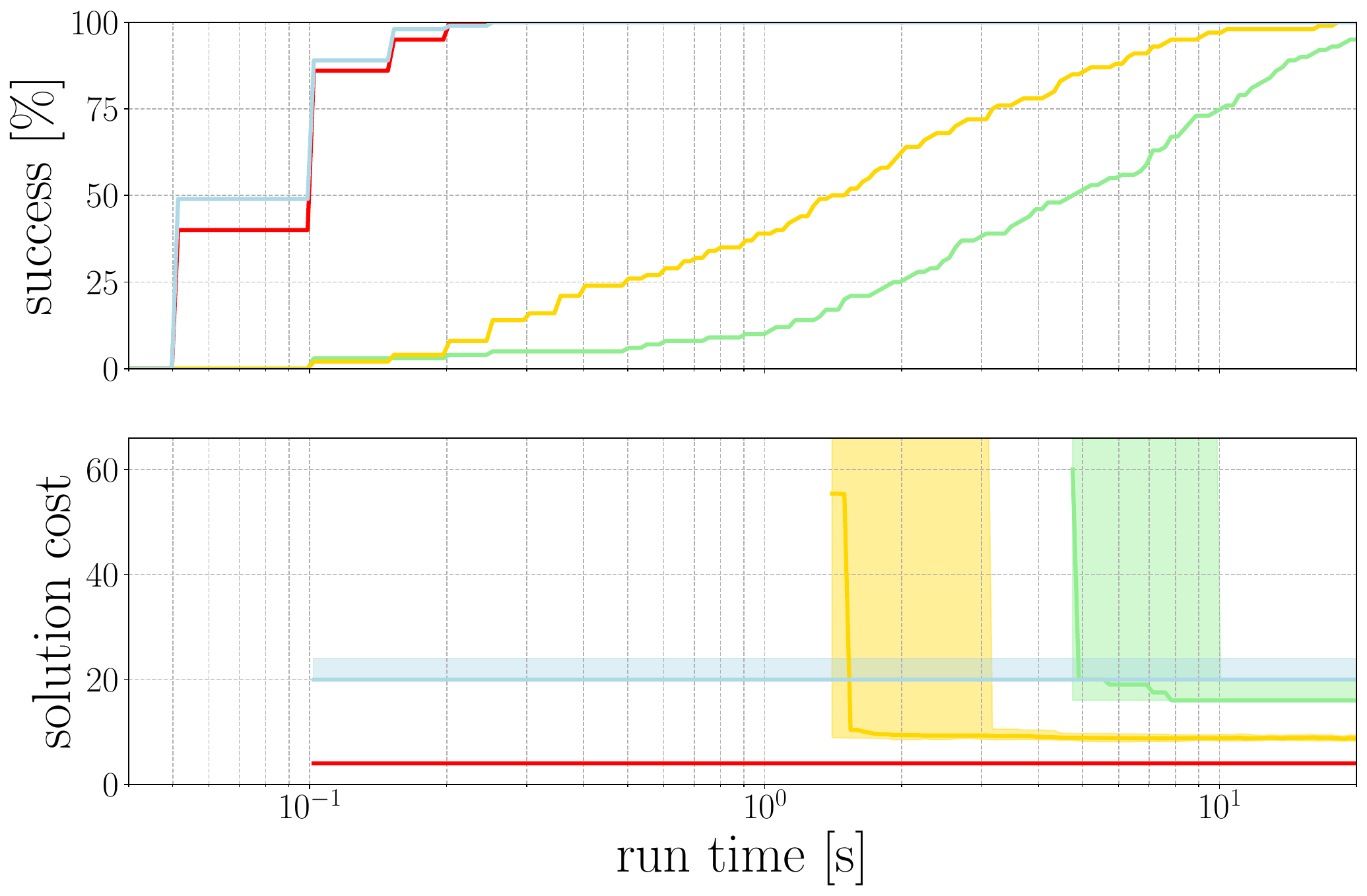}
    \caption*{Continuous Space 4}
  \end{subfigure}\\[1ex]

  \begin{subfigure}[b]{0.24\textwidth}
    \centering
    \includegraphics[width=\textwidth]{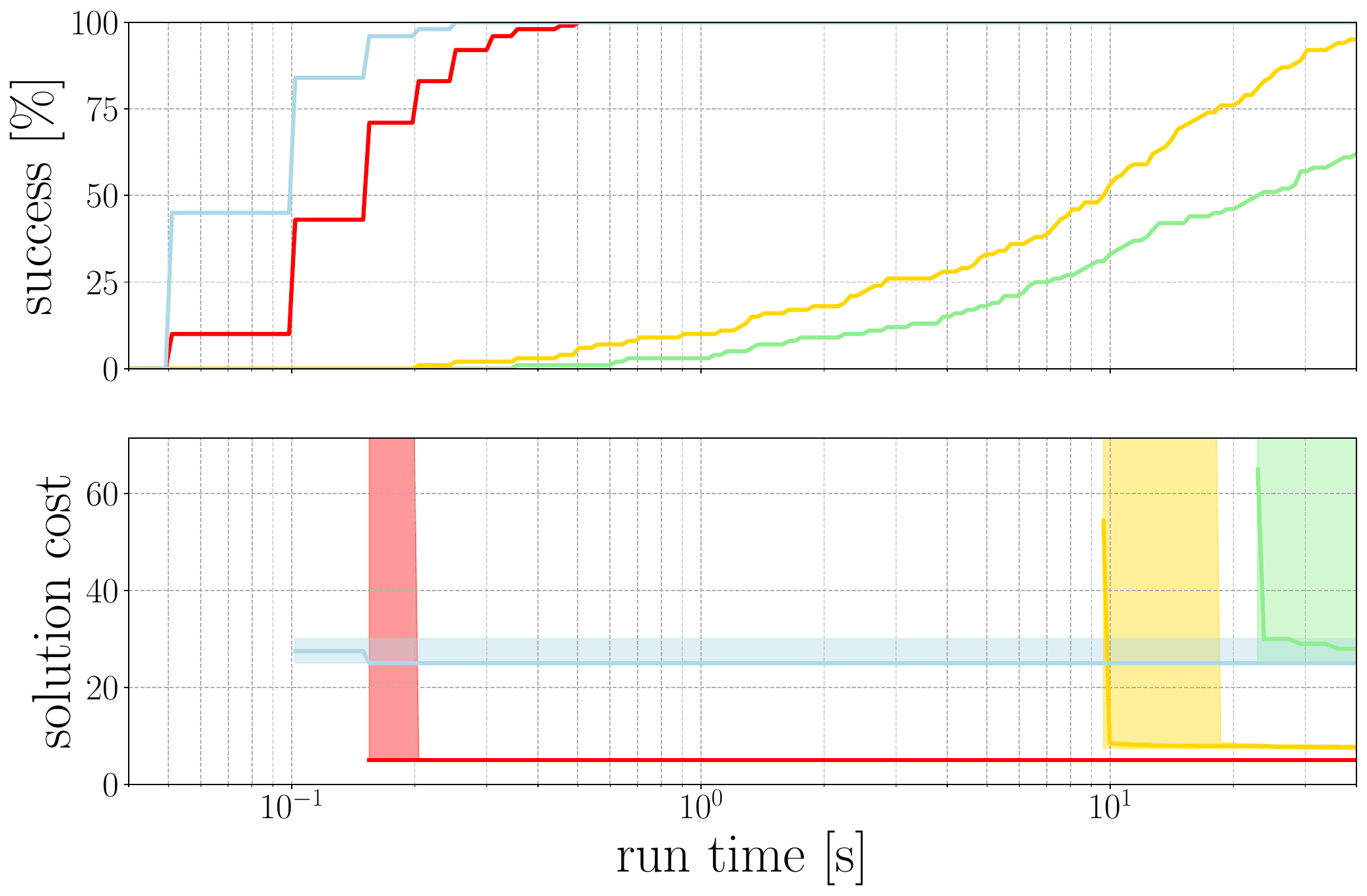}
    \caption*{Grid World 1}
  \end{subfigure}\hfill
  \begin{subfigure}[b]{0.24\textwidth}
    \centering
    \includegraphics[width=\textwidth]{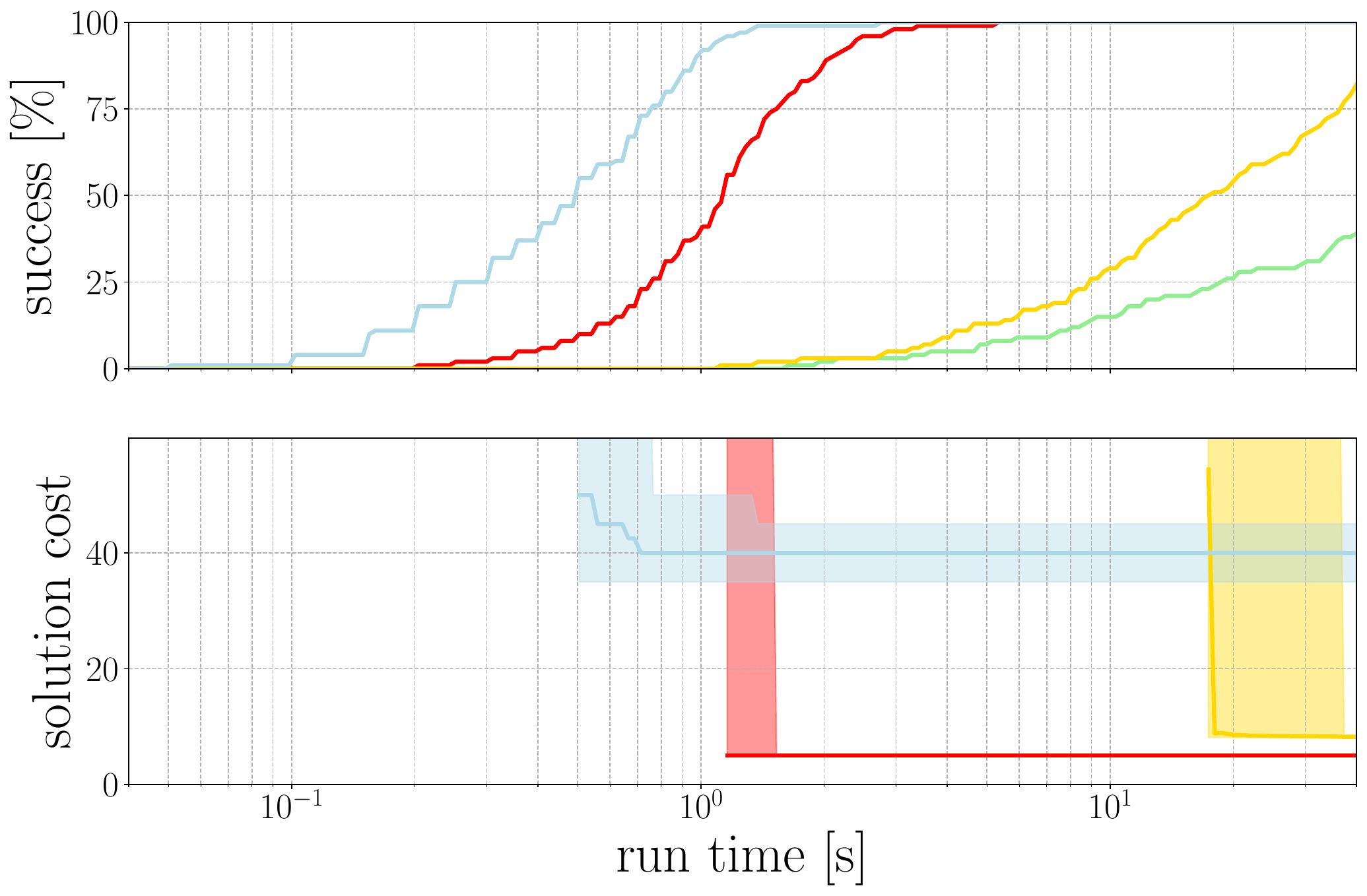}
    \caption*{Grid World 2}
  \end{subfigure}\hfill
  \begin{subfigure}[b]{0.24\textwidth}
    \centering
    \includegraphics[width=\textwidth]{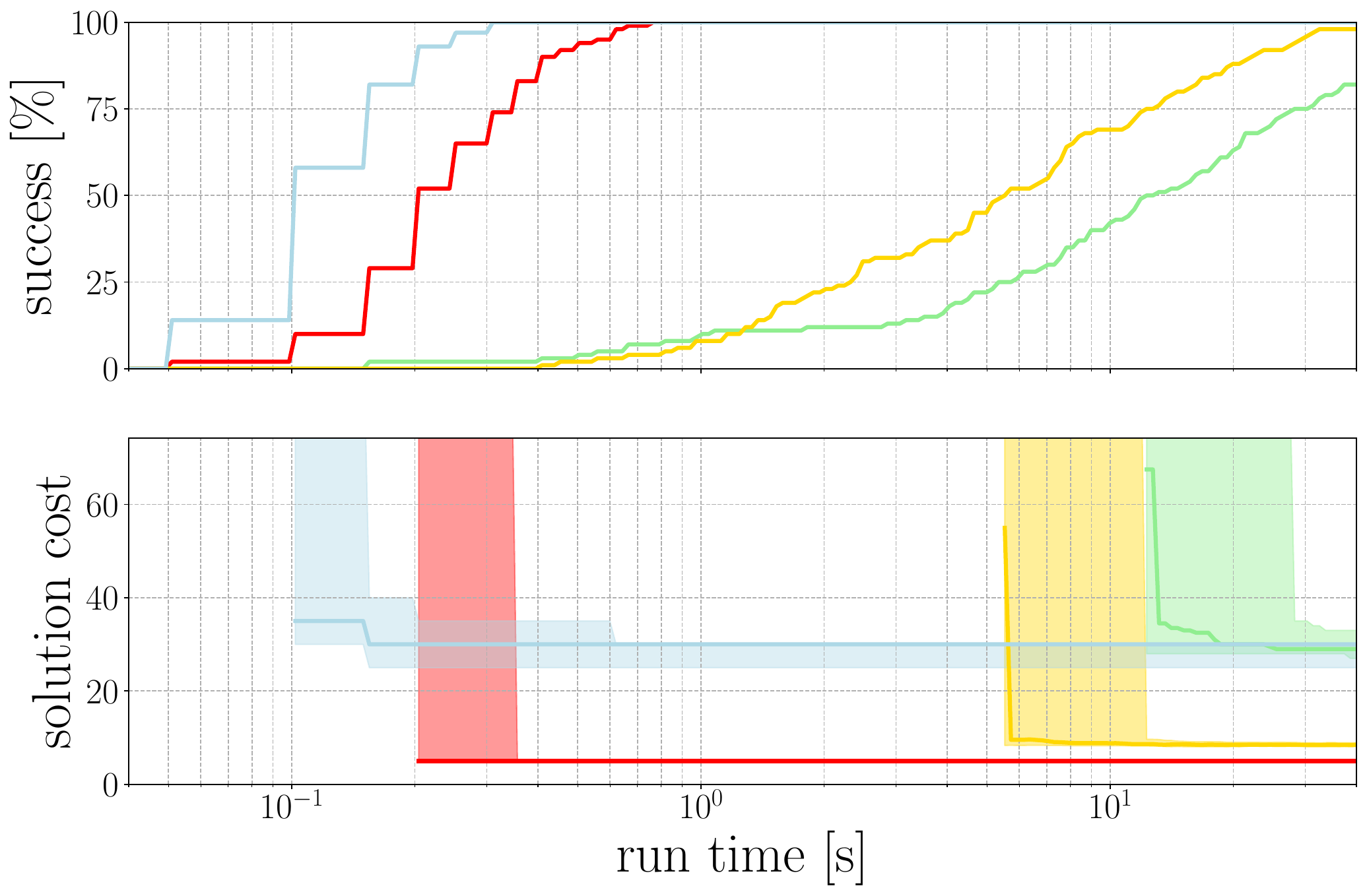}
    \caption*{Grid World 3}
  \end{subfigure}\hfill
  \begin{subfigure}[b]{0.24\textwidth}
    \centering
    \includegraphics[width=\textwidth]{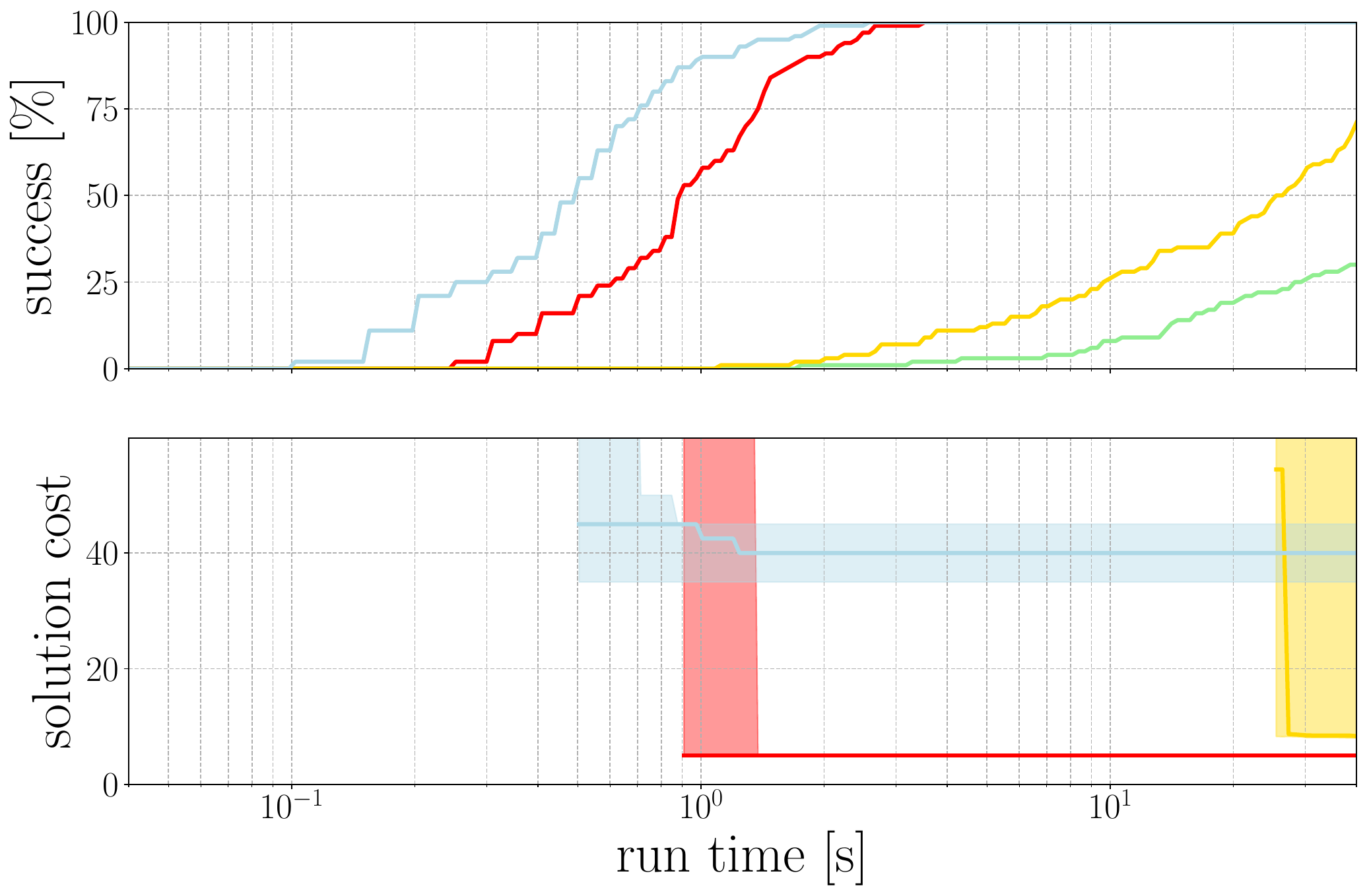}
    \caption*{Grid World 4}
  \end{subfigure}\\[1ex]

  \begin{subfigure}[b]{0.24\textwidth}
    \centering
    \includegraphics[width=\textwidth]{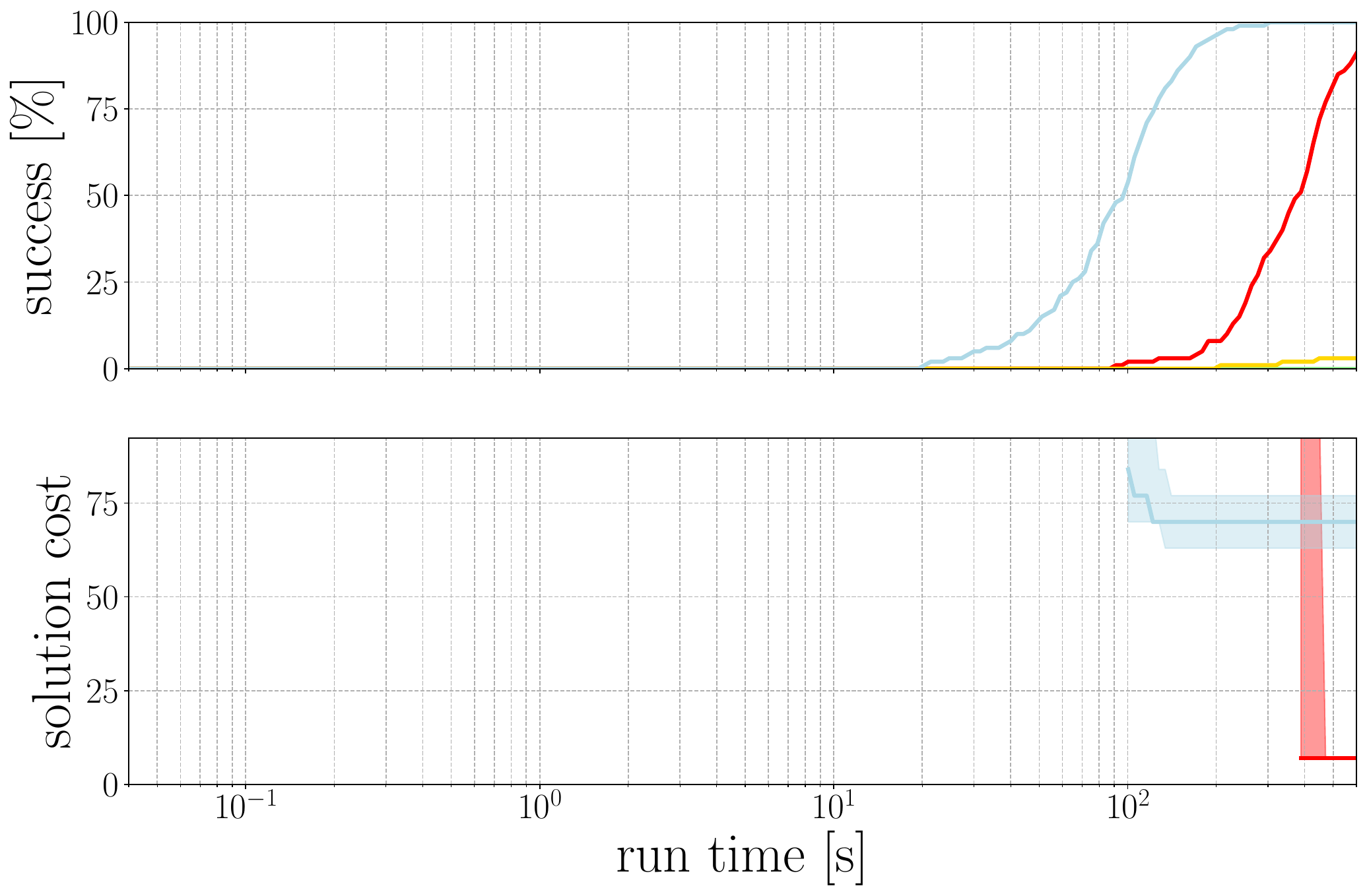}
    \caption*{Lockbox Random 1}
  \end{subfigure}\hfill
  \begin{subfigure}[b]{0.24\textwidth}
    \centering
    \includegraphics[width=\textwidth]{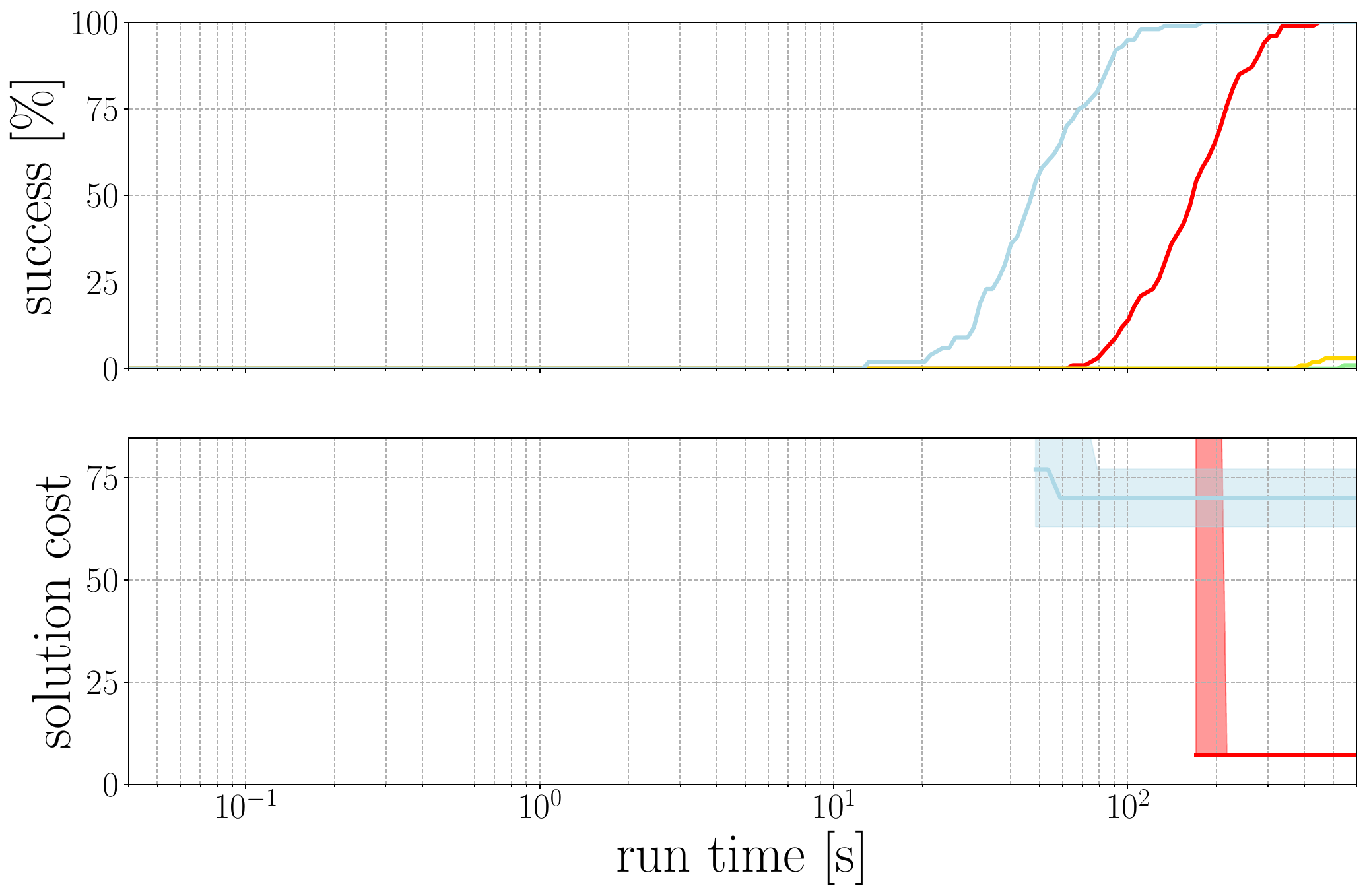}
    \caption*{Lockbox Random 2}
  \end{subfigure}\hfill
  \begin{subfigure}[b]{0.24\textwidth}
    \centering
    \includegraphics[width=\textwidth]{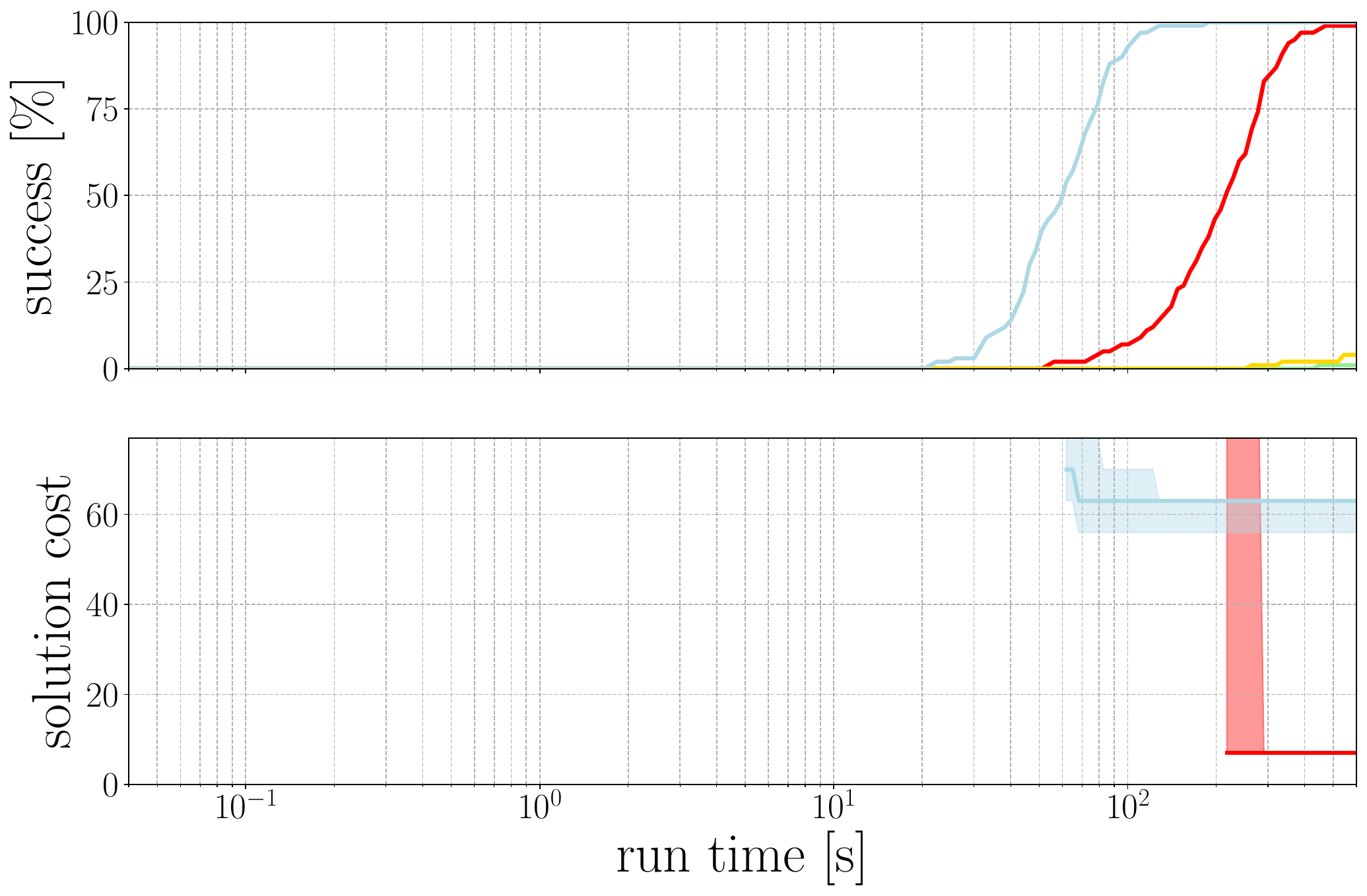}
    \caption*{Lockbox Random 3}
  \end{subfigure}\hfill
  \begin{subfigure}[b]{0.24\textwidth}
    \centering
    \includegraphics[width=\textwidth]{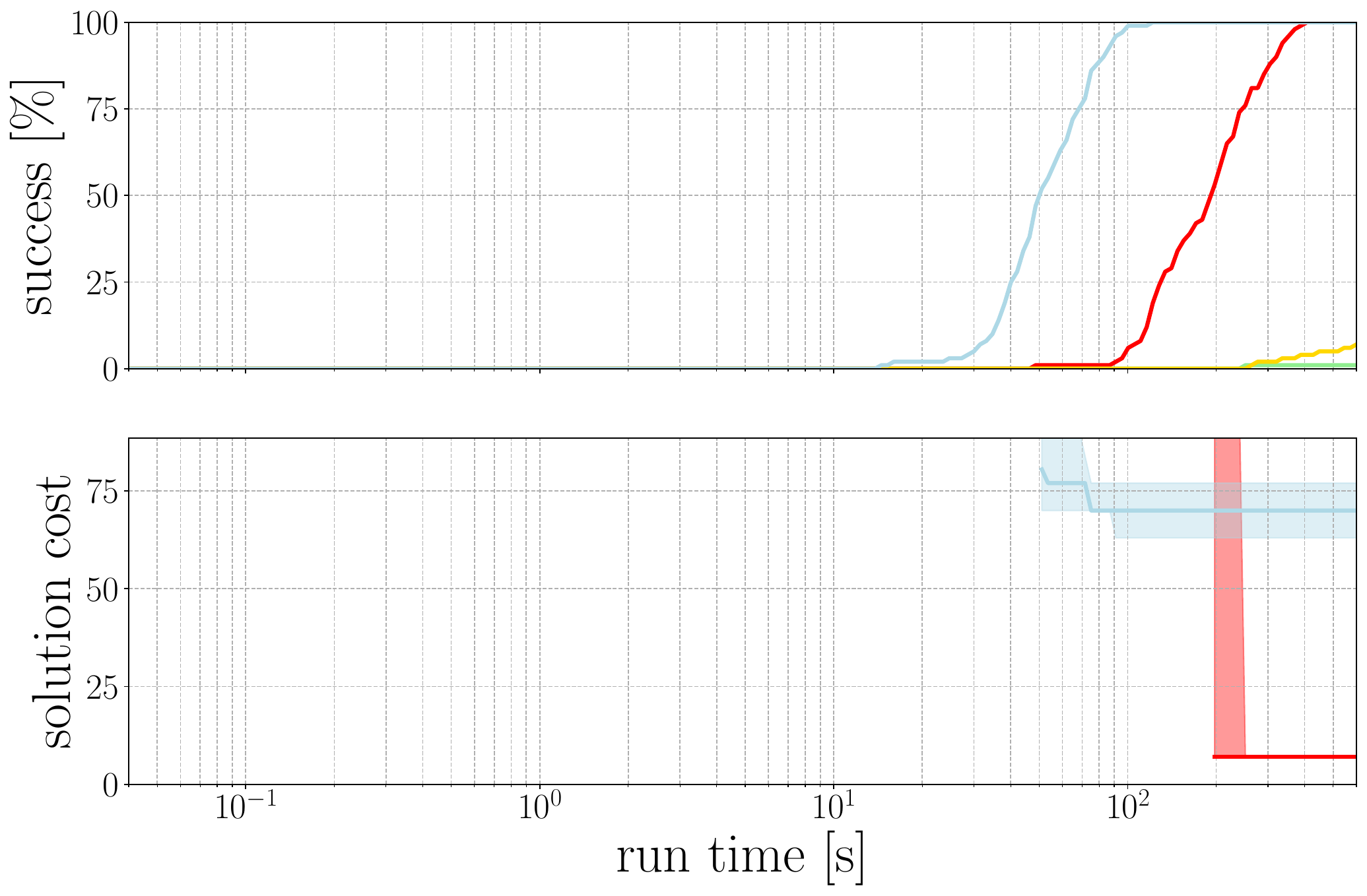}
    \caption*{Lockbox Random 4}
  \end{subfigure}\\[1ex]

  \begin{subfigure}[b]{0.24\textwidth}
    \centering
    \includegraphics[width=\textwidth]{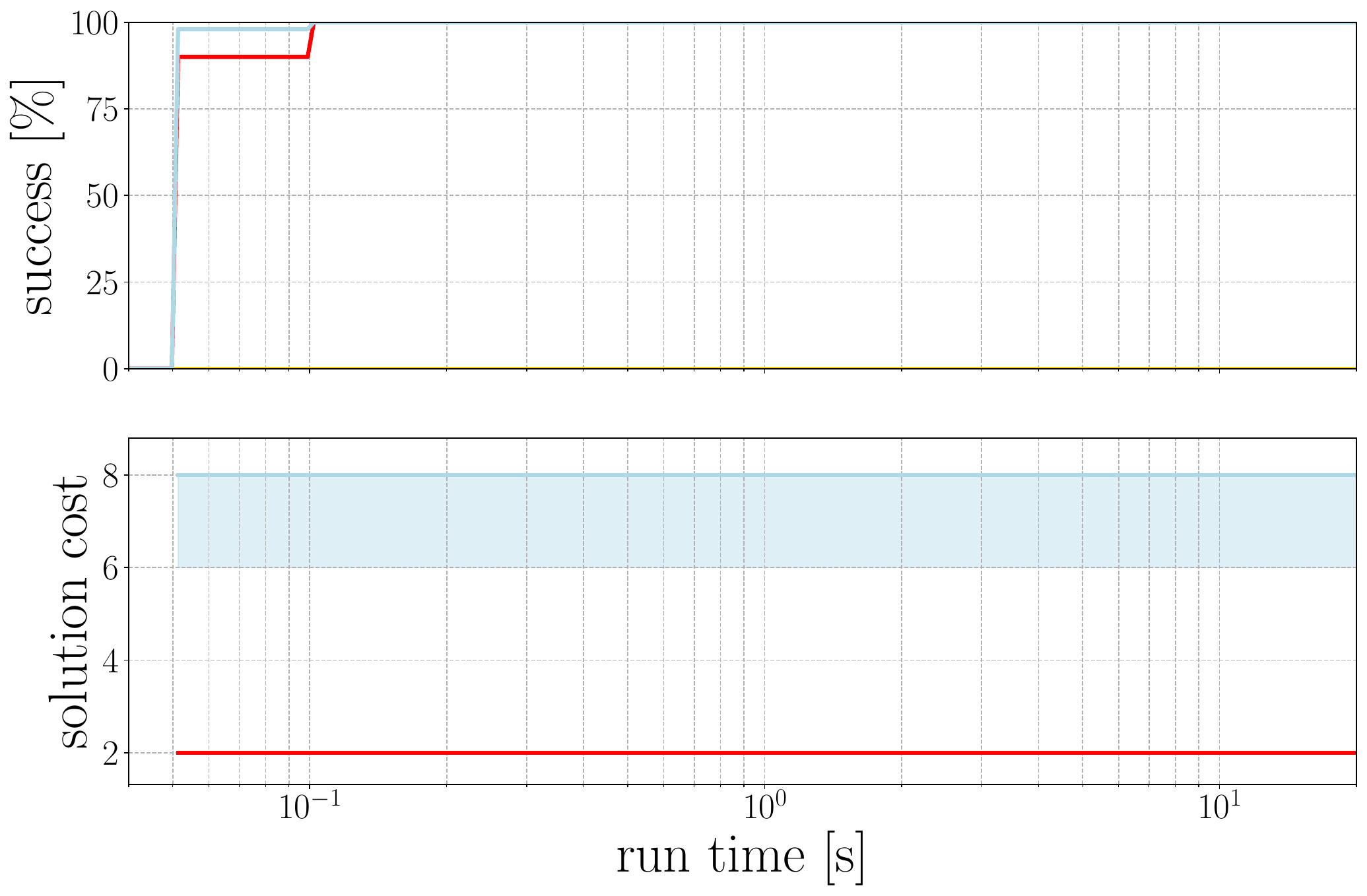}
    \caption*{Move N Times 1}
  \end{subfigure}\hfill
  \begin{subfigure}[b]{0.24\textwidth}
    \centering
    \includegraphics[width=\textwidth]{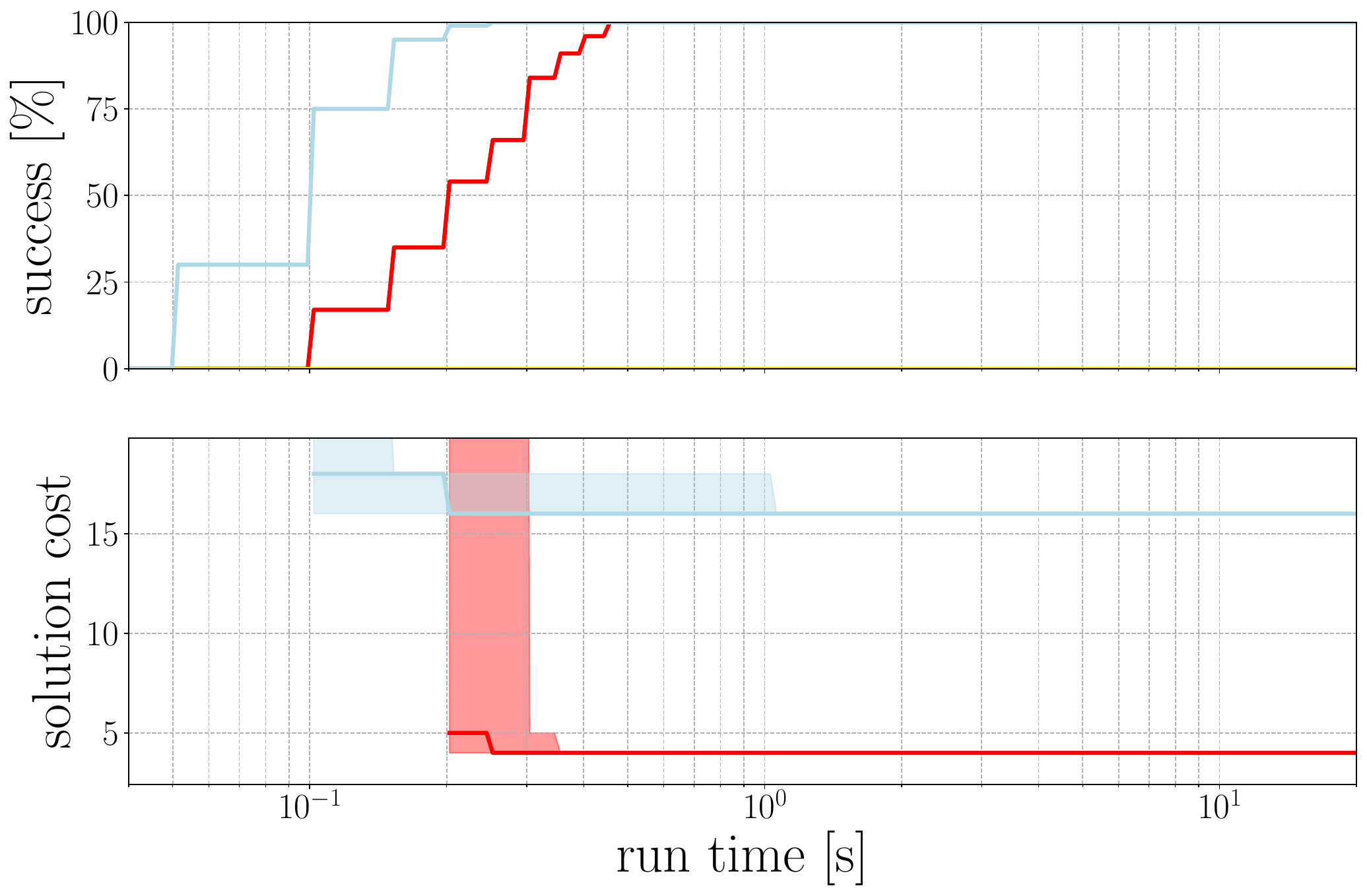}
    \caption*{Move N Times 2}
  \end{subfigure}\hfill
  \begin{subfigure}[b]{0.24\textwidth}
    \centering
    \includegraphics[width=\textwidth]{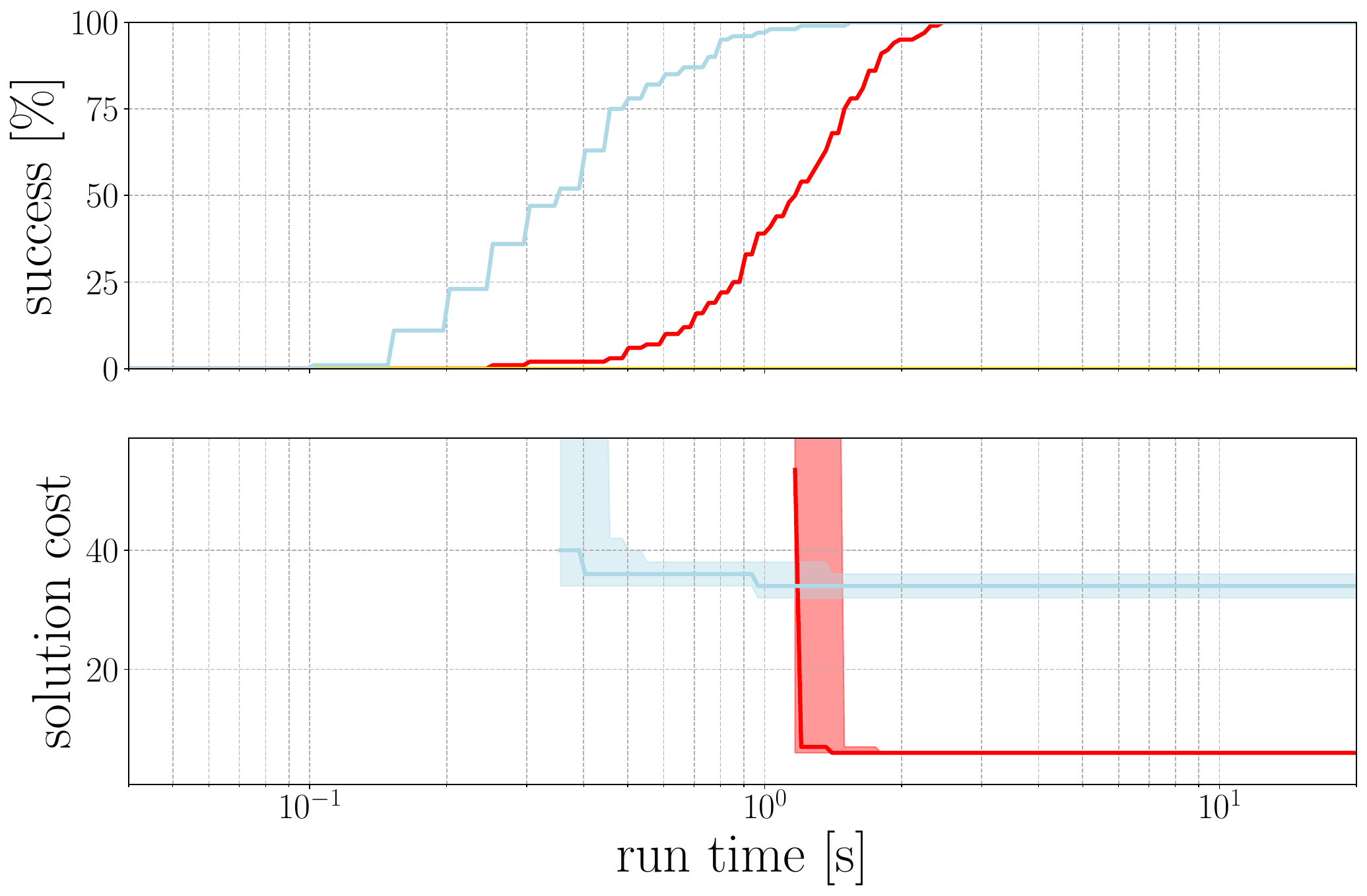}
    \caption*{Move N Times 3}
  \end{subfigure}\hfill
  \begin{subfigure}[b]{0.24\textwidth}
    \centering
    \includegraphics[width=\textwidth]{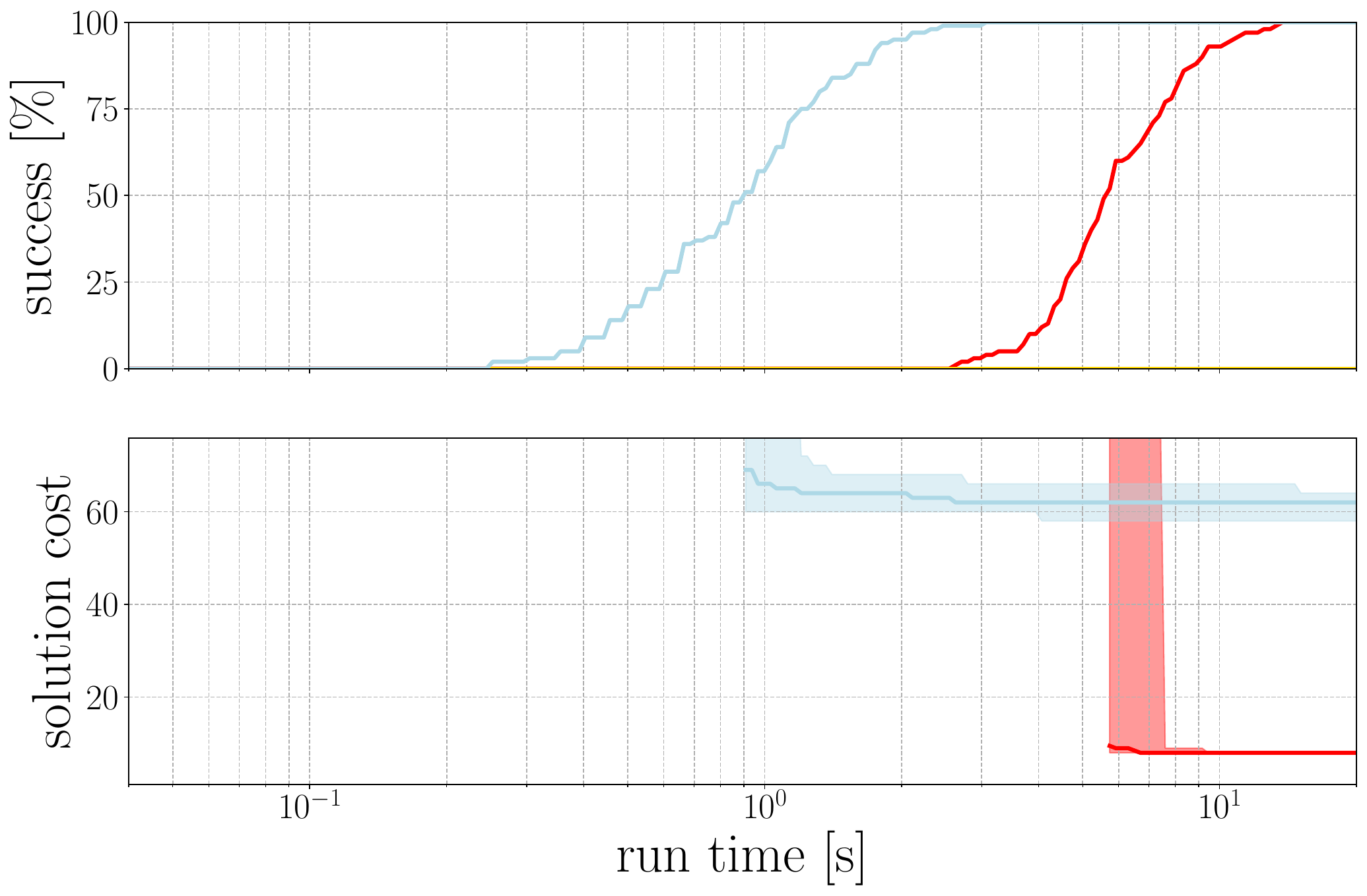}
    \caption*{Move N Times 4}
  \end{subfigure}\\[1ex]

  \begin{subfigure}[b]{0.24\textwidth}
    \centering
    \includegraphics[width=\textwidth]{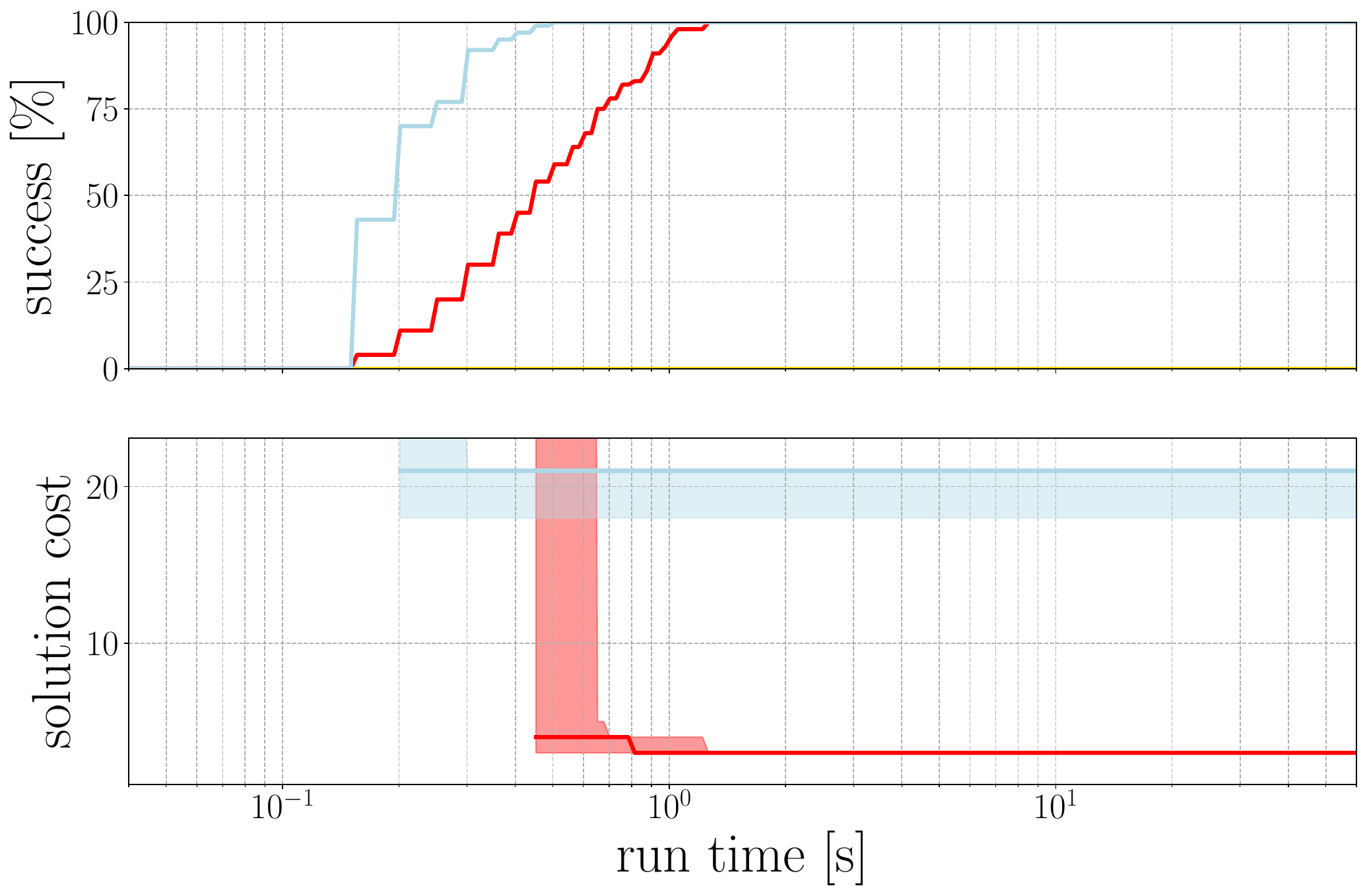}
    \caption*{Rooms 1}
  \end{subfigure}\hfill
  \begin{subfigure}[b]{0.24\textwidth}
    \centering
    \includegraphics[width=\textwidth]{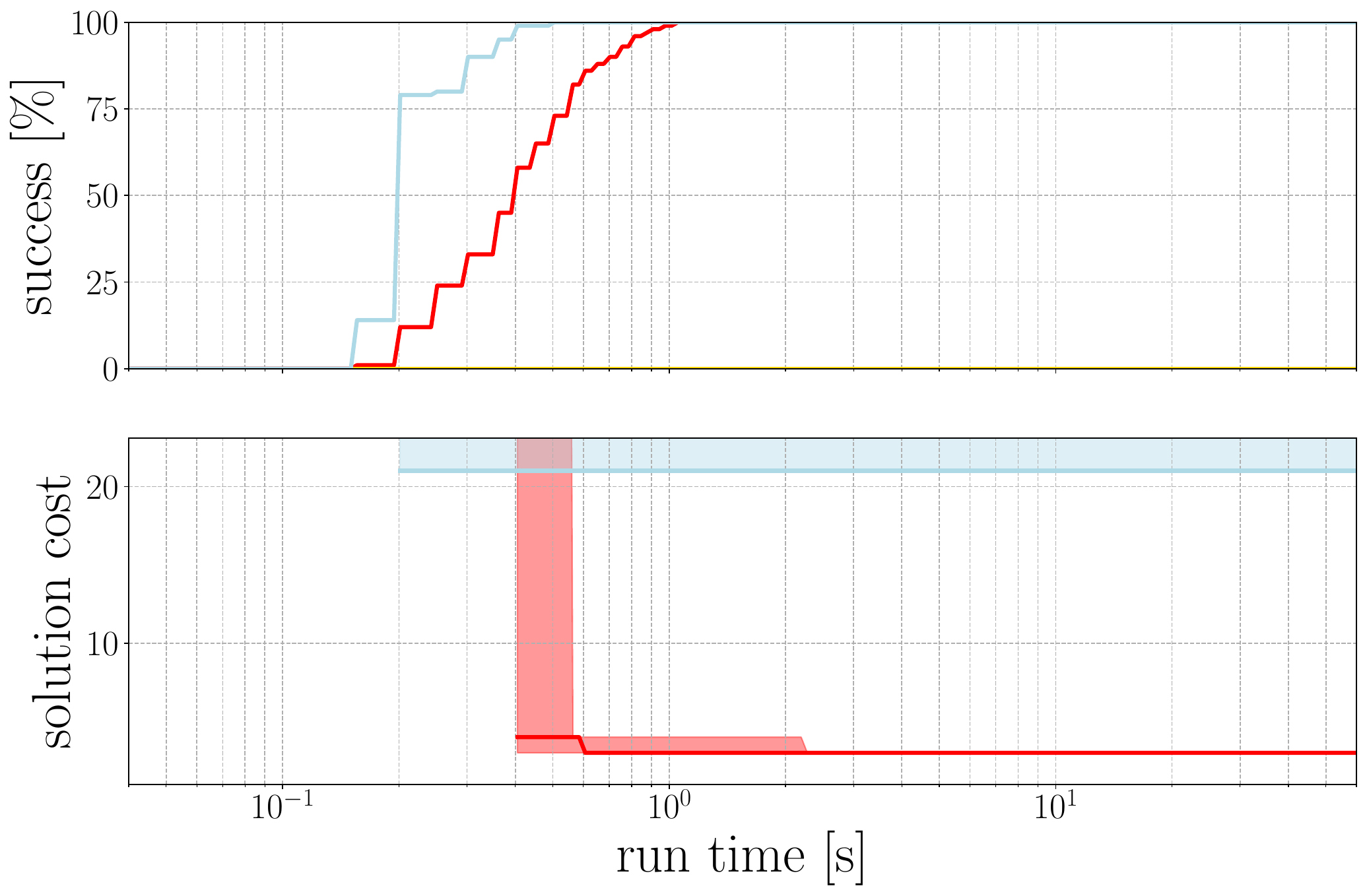}
    \caption*{Rooms 2}
  \end{subfigure}\hfill
  \begin{subfigure}[b]{0.24\textwidth}
    \centering
    \includegraphics[width=\textwidth]{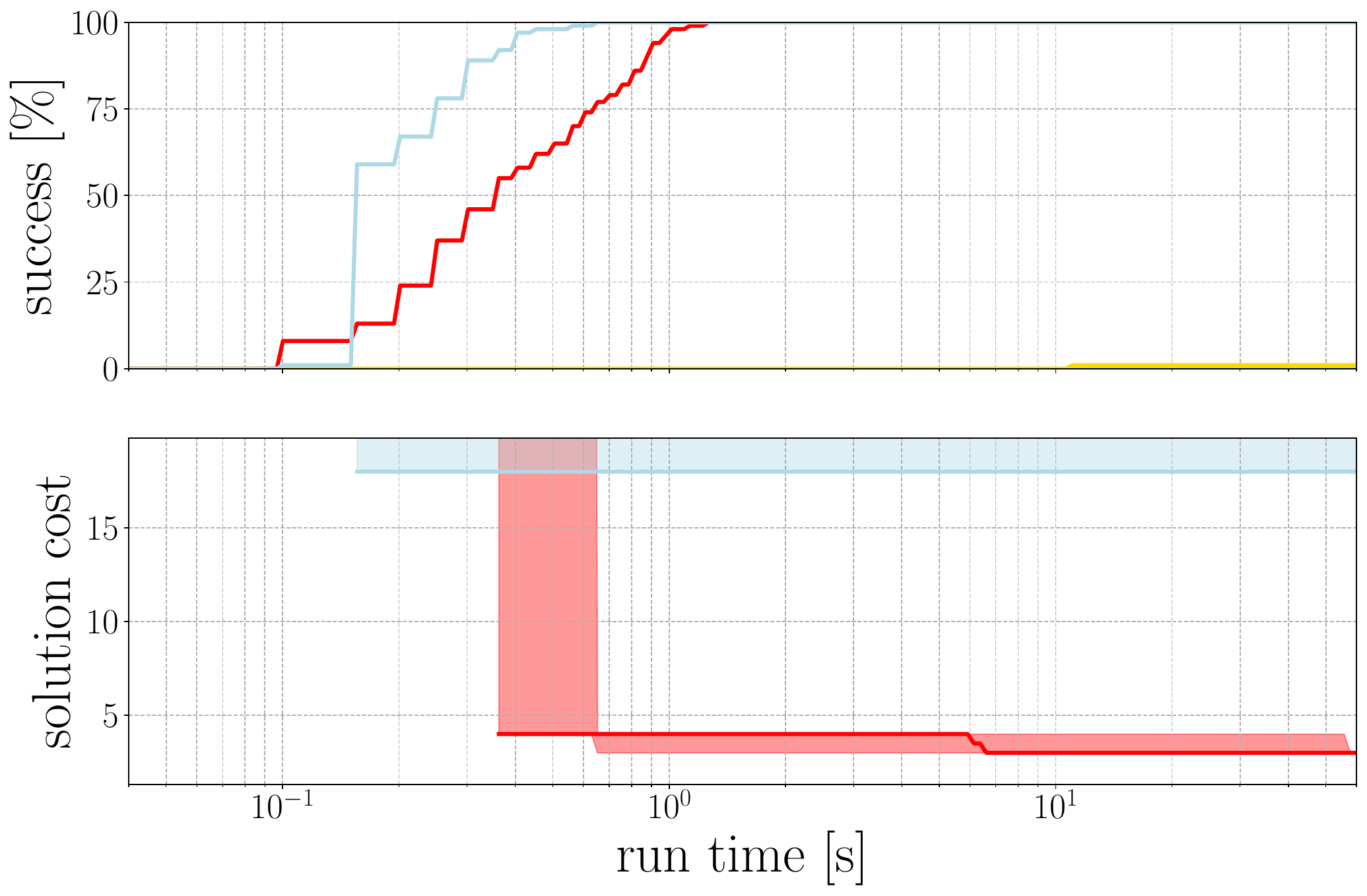}
    \caption*{Rooms 3}
  \end{subfigure}\hfill
  \begin{subfigure}[b]{0.24\textwidth}
    \centering
    \includegraphics[width=\textwidth]{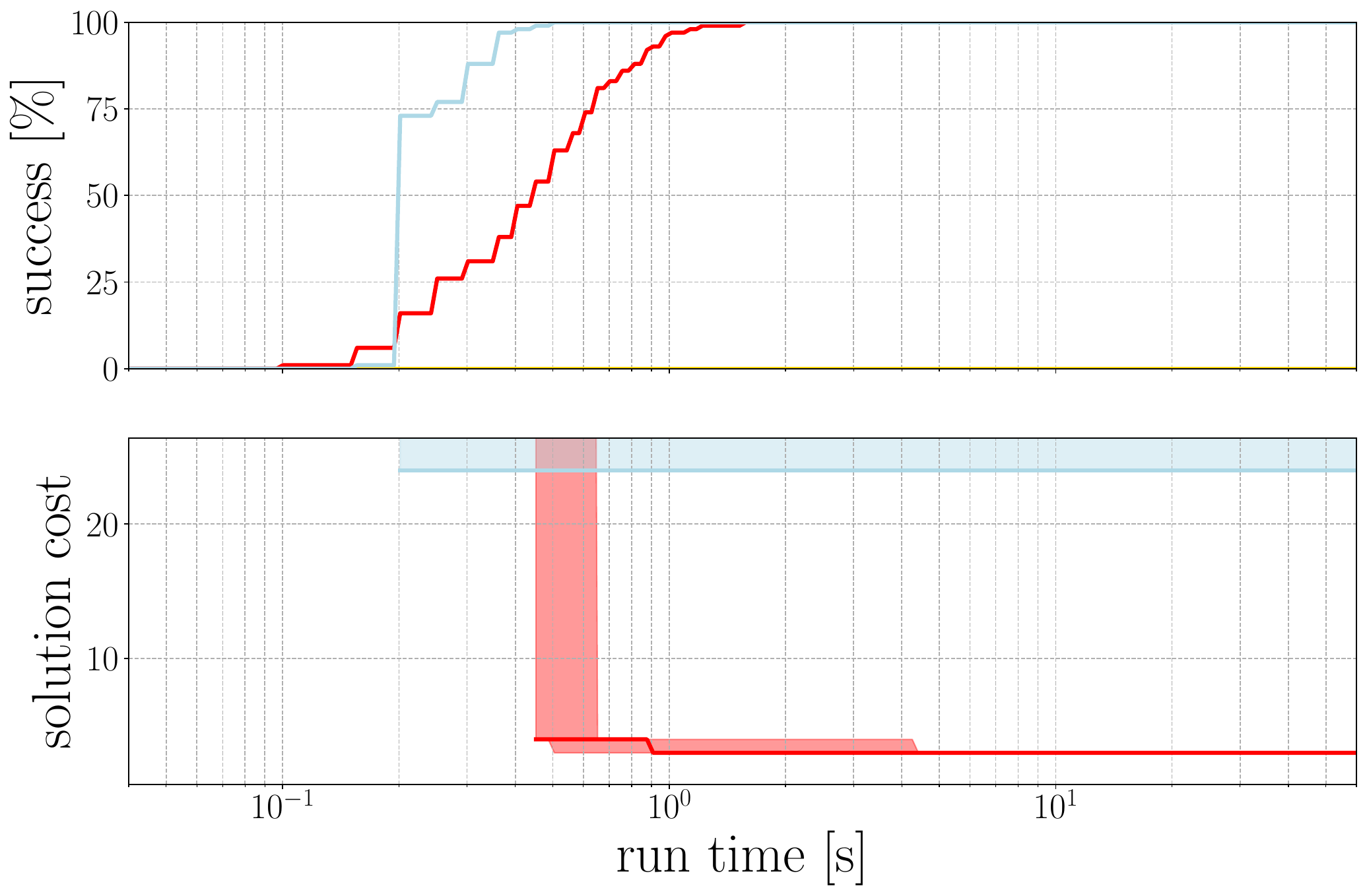}
    \caption*{Rooms 4}
  \end{subfigure}\hfill

  \vskip 1ex
  \includegraphics[width=0.5\textwidth]{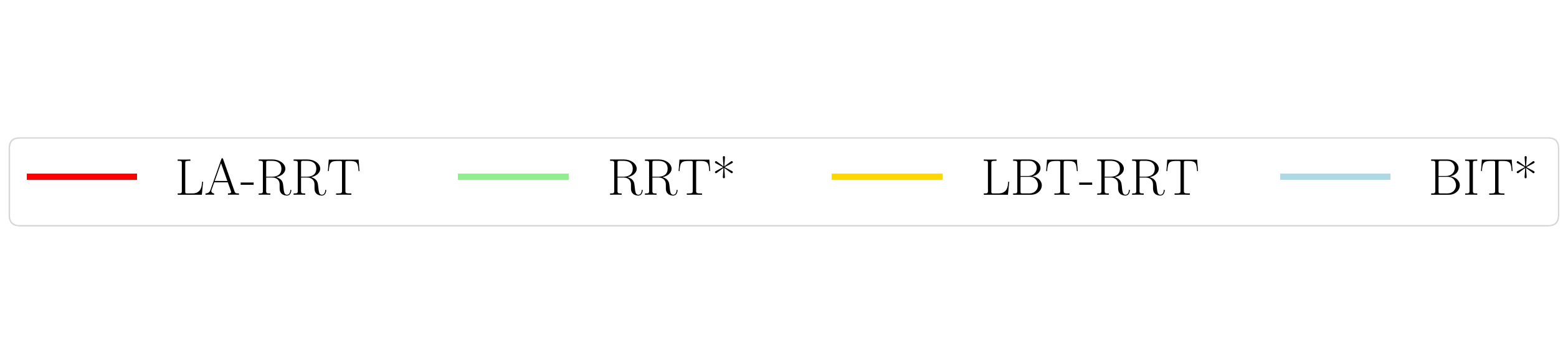}

  \caption{Visualization of benchmark results across six generators, each with four variations, and with the legend below.}
  \label{fig:benchmark}
\end{figure*}

\newcommand{\FrameShift}{1.8cm}
\newcommand{\ImageHeight}{2.4cm}

\begin{figure*}[!htbp]
\centering
\captionsetup[subfigure]{labelformat=empty}

\begin{minipage}[t]{0.02\textwidth}
    \vspace*{-\FrameShift}\rotatebox{90}{\parbox{1.5cm}{\raggedleft Frame 1}}
\end{minipage}\hfill
\begin{subfigure}[t]{0.15\textwidth}
    \centering
    \includegraphics[width=\linewidth,height=\ImageHeight]{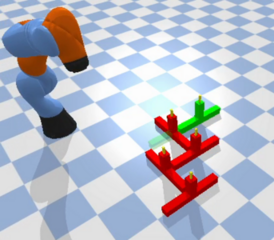}
\end{subfigure}\hfill
\begin{subfigure}[t]{0.15\textwidth}
    \centering
    \includegraphics[width=\linewidth,height=\ImageHeight]{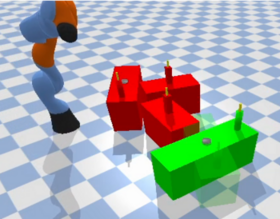}
\end{subfigure}\hfill
\begin{subfigure}[t]{0.15\textwidth}
    \centering
    \includegraphics[width=\linewidth,height=\ImageHeight]{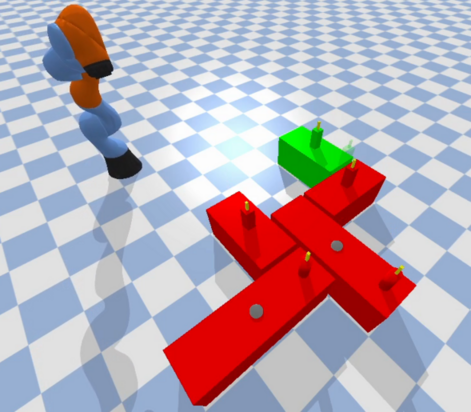}
\end{subfigure}\hfill
\begin{subfigure}[t]{0.15\textwidth}
    \centering
    \includegraphics[width=\linewidth,height=\ImageHeight]{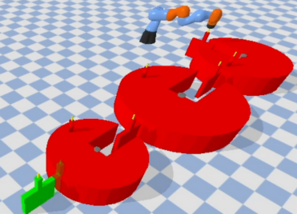}
\end{subfigure}\hfill
\begin{subfigure}[t]{0.15\textwidth}
    \centering
    \includegraphics[width=\linewidth,height=\ImageHeight]{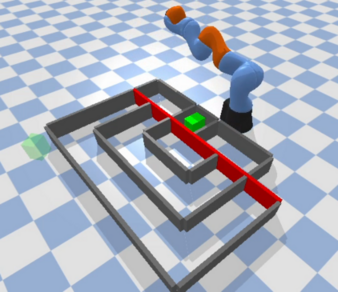}
\end{subfigure}\hfill
\begin{subfigure}[t]{0.15\textwidth}
    \centering
    \includegraphics[width=\linewidth,height=\ImageHeight]{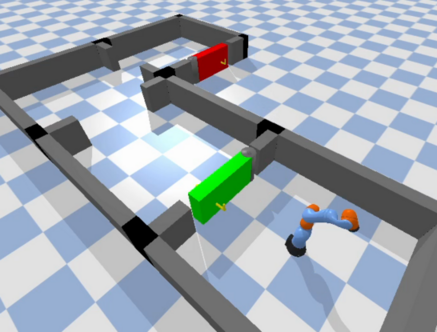}
\end{subfigure}

\vspace{2mm}

\begin{minipage}[t]{0.02\textwidth}
    \vspace*{-\FrameShift}\rotatebox{90}{\parbox{1.5cm}{\raggedleft Frame 2}}
\end{minipage}\hfill
\begin{subfigure}[t]{0.15\textwidth}
    \centering
    \includegraphics[width=\linewidth,height=\ImageHeight]{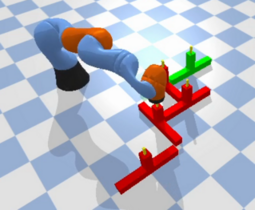}
\end{subfigure}\hfill
\begin{subfigure}[t]{0.15\textwidth}
    \centering
    \includegraphics[width=\linewidth,height=\ImageHeight]{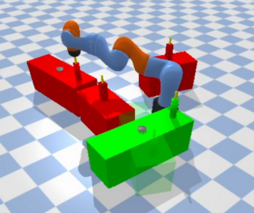}
\end{subfigure}\hfill
\begin{subfigure}[t]{0.15\textwidth}
    \centering
    \includegraphics[width=\linewidth,height=\ImageHeight]{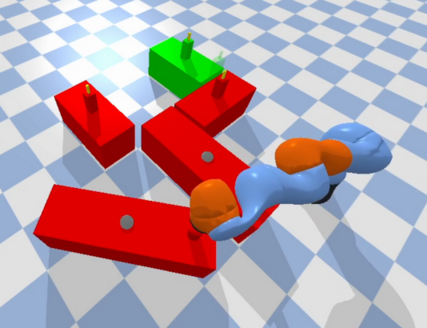}
\end{subfigure}\hfill
\begin{subfigure}[t]{0.15\textwidth}
    \centering
    \includegraphics[width=\linewidth,height=\ImageHeight]{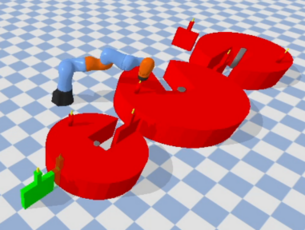}
\end{subfigure}\hfill
\begin{subfigure}[t]{0.15\textwidth}
    \centering
    \includegraphics[width=\linewidth,height=\ImageHeight]{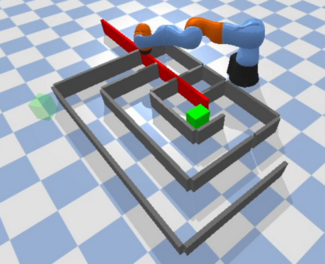}
\end{subfigure}\hfill
\begin{subfigure}[t]{0.15\textwidth}
    \centering
    \includegraphics[width=\linewidth,height=\ImageHeight]{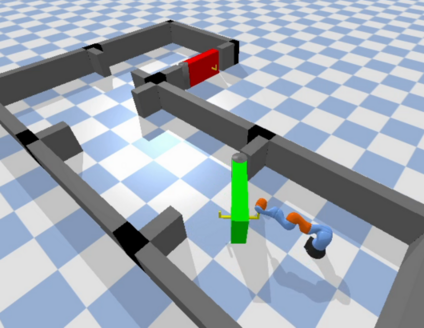}
\end{subfigure}

\vspace{2mm}

\begin{minipage}[t]{0.02\textwidth}
    \vspace*{-\FrameShift}\rotatebox{90}{\parbox{1.5cm}{\raggedleft Frame 3}}
\end{minipage}\hfill
\begin{subfigure}[t]{0.15\textwidth}
    \centering
    \includegraphics[width=\linewidth,height=\ImageHeight]{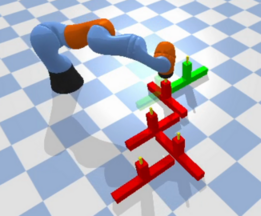}
\end{subfigure}\hfill
\begin{subfigure}[t]{0.15\textwidth}
    \centering
    \includegraphics[width=\linewidth,height=\ImageHeight]{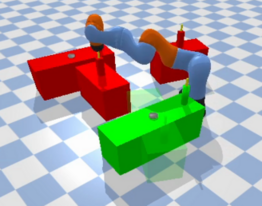}
\end{subfigure}\hfill
\begin{subfigure}[t]{0.15\textwidth}
    \centering
    \includegraphics[width=\linewidth,height=\ImageHeight]{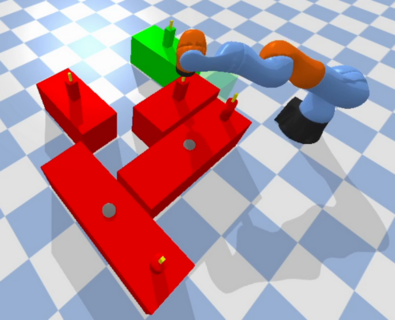}
\end{subfigure}\hfill
\begin{subfigure}[t]{0.15\textwidth}
    \centering
    \includegraphics[width=\linewidth,height=\ImageHeight]{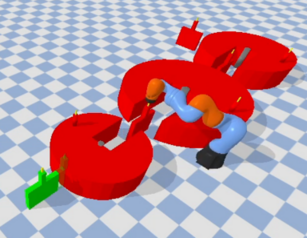}
\end{subfigure}\hfill
\begin{subfigure}[t]{0.15\textwidth}
    \centering
    \includegraphics[width=\linewidth,height=\ImageHeight]{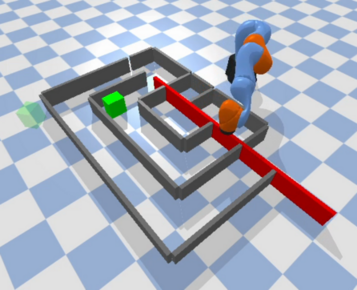}
\end{subfigure}\hfill
\begin{subfigure}[t]{0.15\textwidth}
    \centering
    \includegraphics[width=\linewidth,height=\ImageHeight]{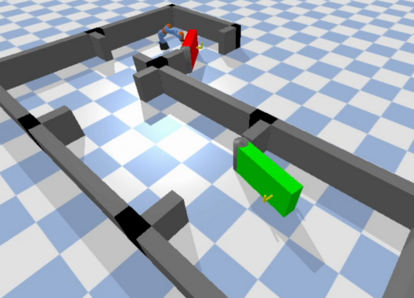}
\end{subfigure}

\vspace{2mm}

\begin{minipage}[t]{0.02\textwidth}
    \vspace*{-\FrameShift}\rotatebox{90}{\parbox{1.5cm}{\raggedleft Frame 4}}
\end{minipage}\hfill
\begin{subfigure}[t]{0.15\textwidth}
    \centering
    \includegraphics[width=\linewidth,height=\ImageHeight]{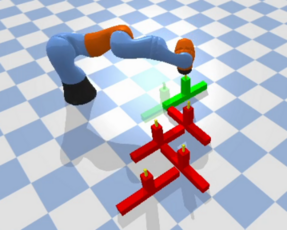}
\end{subfigure}\hfill
\begin{subfigure}[t]{0.15\textwidth}
    \centering
    \includegraphics[width=\linewidth,height=\ImageHeight]{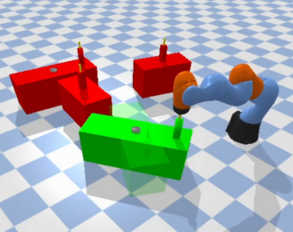}
\end{subfigure}\hfill
\begin{subfigure}[t]{0.15\textwidth}
    \centering
    \includegraphics[width=\linewidth,height=\ImageHeight]{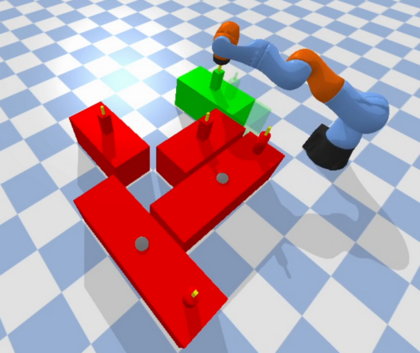}
\end{subfigure}\hfill
\begin{subfigure}[t]{0.15\textwidth}
    \centering
    \includegraphics[width=\linewidth,height=\ImageHeight]{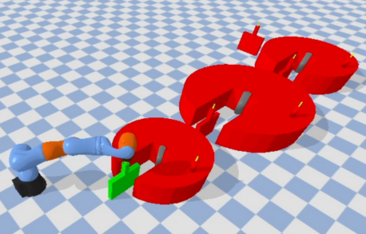}
\end{subfigure}\hfill
\begin{subfigure}[t]{0.15\textwidth}
    \centering
    \includegraphics[width=\linewidth,height=\ImageHeight]{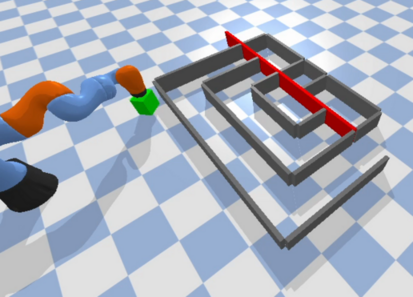}
\end{subfigure}\hfill
\begin{subfigure}[t]{0.15\textwidth}
    \centering
    \includegraphics[width=\linewidth,height=\ImageHeight]{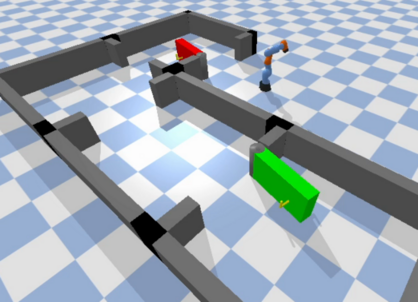}
\end{subfigure}

\vspace{2mm}
\begin{minipage}[t]{0.02\textwidth}
    \centering
    \phantom{.}
\end{minipage}\hfill
\begin{minipage}[t]{0.15\textwidth}
    \centering
    \small Simple Slider
\end{minipage}\hfill
\begin{minipage}[t]{0.15\textwidth}
    \centering
    \small Continuous Space
\end{minipage}\hfill
\begin{minipage}[t]{0.15\textwidth}
    \centering
    \small Grid World
\end{minipage}\hfill
\begin{minipage}[t]{0.15\textwidth}
    \centering
    \small Lockbox Random
\end{minipage}\hfill
\begin{minipage}[t]{0.15\textwidth}
    \centering
    \small Move N Times
\end{minipage}\hfill
\begin{minipage}[t]{0.15\textwidth}
    \centering
    \small Rooms
\end{minipage}

\caption{Visualization of manipulation tasks across all six environments. The manipulation task for the mobile manipulator robot is to move the green object to its goal position (transparent green), while red objects block the green object and have to be moved out of the way (interlocking dependencies). The \texttt{Rooms} environment is different in that the robot has to move out of the room itself, but is blocked by all objects (doors).
\label{fig:manipulation_environments}}
\end{figure*}

\section{Demonstration}

The demonstration is divided into three parts. 
In the first part, we show how our generators can generate different puzzle variations.
In the second part, we use different sampling-based motion planning algorithms to benchmark solving those puzzles. 
Finally, we execute each puzzle with a KUKA LBR iiwa robot by using the best solution found by the benchmark planners. 

Our code is available on GitHub\footnote{\url{https://github.com/ljdross/puzzle-generator}}. 
All experiments were conducted on a workstation PC with an Intel Core i7-6700 3.4 Ghz quad-core processor, 16 GB memory, running the Ubuntu 18.04 LTS operating system.

\subsection{Puzzle Generation}

We show that our generators can generate unique physical manipulation puzzles using our framework. 
To generate the puzzles, we ran each generator and generate the first four seeds for each puzzle. The runtimes to generate them are Simple Sliders ($4$ seconds), Grid World ($5$ seconds), Continuous Space ($83$ seconds), Lockbox Random ($13$ seconds), Move N Times ($5$ seconds), and Rooms ($17$ seconds). This generates $24$ unique physical puzzles as shown in Fig.~\ref{fig:puzzle_environments}. 

\subsection{Puzzle Benchmarking}

To show that the generated puzzles are solvable, we use four sampling-based planning algorithms on all 24 puzzles. The four planning algorithms are LA-RRT~\cite{Bayraktar2023RAL}, RRT*~\cite{karaman2011sampling}, LBT-RRT~\cite{salzman2016asymptotically}, and BIT*~\cite{gammell2020batch}. 
Note that we are using optimal sampling-based planners~\cite{gammell2021asymptotically}, since we want to optimize the total action cost, which is the total number of actions (individual joint moves) required to solve the puzzle. Each planner is run on each puzzle whereby we repeat the experiments $100$ times per planner with the following time limits: Simple Sliders ($40$ seconds), Grid World ($40$ seconds), Continuous Space ($20$ seconds), Lockbox Random ($600$ seconds), Move N Times ($20$ seconds), and Rooms ($60$ seconds). We collect both the time to find the first feasible solution and track the cost over time.

The averaged results are shown in Fig.~\ref{fig:benchmark}. It can be seen that LA-RRT is able to solve $23$ out of $24$ scenarios with a success rate of $100\%$. The only outlier is Lockbox Random 1, where LA-RRT only reaches $95\%$ success rate at the time limit. BIT* is able to solve $24$ out of $24$ scenarios with a success rate of $100\%$. The two remaining planners LBT-RRT and RRT* can solve $7$ out of $24$ and $6$ out of $24$ with $100\%$. In terms of optimality, LA-RRT reaches the lowest cost of all planners consistently over all $24$ scenarios. This is expected, since LA-RRT is specifically designed to solve manipulation puzzles with a minimal number of actions.

\subsection{Puzzle Manipulation}

By using the method described in Sec.~\ref{sec:benchmark-and-manip}, we can show that all $24$ puzzles can eventually be executed using a mobile manipulator robot. The results are depicted in Fig.~\ref{fig:manipulation_environments}, where we show snapshots of how the robot is manipulating each puzzle. 
For each scenario, we show four frames, whereby Frame $1$ depicts the start configuration of the puzzle. Frame $2$ and Frame $3$ show intermediate manipulation steps, where the robot moves some of the dependent obstacles out of the way. Eventually, Frame $4$ shows the last action where the robot moves the green object (the only one with a dedicated goal configuration) towards its goal position. 
The robot used is a KUKA LBR iiwa on a mobile manipulator base, having three degrees of freedom (dof) for the base and $7$-dof for the manipulator arm. 
Full videos for all $24$ scenarios are available\footnote{ \href{https://sites.google.com/view/robot-manipulation-puzzles}{sites.google.com/view/robot-manipulation-puzzles}}.
The complete execution and planning time for the scenarios is between $4$s (Simple Sliders 1) and $51$s (Move N Times 4).
\section{Conclusion}

We proposed to leverage procedural content generation (PCG) to generate synthetic datasets of manipulation puzzles suitable for learning-based approaches that involve physical exploration.

Compared with previous works, such as ProcTHOR \cite{deitke2022procthor} and MotionBenchMaker \cite{Chamzas2022RAL}, which also leverage PCG to generate manipulation problems, our method uses PCG specifically to generate interlocking object dependencies to create physical manipulation puzzles.
The additional aspect of exploring and reasoning about the physical constraints of environments makes our methods suitable for learning-based TAMP~\cite{zhao2024survey}.

While we have shown that our method is able to generate unique manipulation
puzzles, we still face some limitations.

\begin{itemize}

    \item \textbf{Narrow Passages} 
      Our current method struggles to generate environments containing narrow
      passages between objects, because motion planning methods often cannot handle them efficiently~\cite{Orthey2021TRO}. We could solve this by using planners combining sampling-based and optimization-based frameworks~\cite{Kamat2022IROS}.

    \item \textbf{Tree-Dependencies} 
      Our current framework generates environments with sequential (chain-like) dependencies between objects.
      However, puzzles often require tree-like dependency structures, where multiple objects independently block a common target, which we aim to generate in future work using a backtracking algorithm.


    \item \textbf{Online Learning} 
      Our current framework generates data which can then be used for learning. However, we envision that our framework is used in an online manner, where the difficulty of environments is adjusted based on the progress of the robot~\cite{gisslen2021adversarial}.

    \item \textbf{Baseline Comparisons}
      Our evaluation demonstrates solvability and manipulatability of the generated puzzles, but does not yet include comparisons against other generation frameworks or ablation studies. A systematic comparison with manually curated benchmarks would provide stronger evidence for the utility of procedurally generated puzzles.

    \item \textbf{Real-World Validation}
      All experiments are conducted in simulation. Transferring the generated puzzles to a physical robot would validate that the puzzles and solutions transfer to a real robot.

\end{itemize}

Despite limitations, our proposed method is able to reliably generate synthetic
datasets of unique manipulation puzzles in seconds.
The difficulty and solvability of the 24 puzzle samples is verified by extensive benchmarking and manipulation simulations.
Overall, our method yields good performance in terms of speed of generation ($4$s to $83$s), speed of solvability ($1$s to $300$s), and quality of
generated puzzles (all are solvable using a realistic mobile manipulator). We
therefore believe that this framework is a useful tool to create large datasets for benchmarking and
training of learning-based manipulation robots~\cite{kroemer2021review,firoozi2025foundation}.

\bibliographystyle{IEEEtran}
{
\balance
\small
\bibliography{bib/general}

@article{zhao2024survey,
title={A Survey of Optimization-Based Task and Motion Planning: From Classical to Learning Approaches},
author={Zhao, Zhigen and Cheng, Shuo and Ding, Yan and others},
journal={IEEE/ASME Transactions on Mechatronics},
year={2024},
publisher={IEEE}
}

@article{kroemer2021review,
title={A Review of Robot Learning for Manipulation: Challenges, Representations, and Algorithms},
author={Kroemer, Oliver and Niekum, Scott and Konidaris, George},
journal={Journal of Machine Learning Research},
volume={22},
number={30},
pages={1--82},
year={2021}
}

@inproceedings{gisslen2021adversarial,
title={Adversarial Reinforcement Learning for Procedural Content Generation},
author={Gissl{\'e}n, Linus and Eakins, Andy and Gordillo, Camilo and Bergdahl, Joakim and Tollmar, Konrad},
booktitle={Proceedings of the IEEE Conference on Games},
pages={1--8},
year={2021},
organization={IEEE}
}

@book{Sutton2018reinforcement,
title={Reinforcement Learning: An Introduction},
author={Sutton, Richard S and Barto, Andrew G},
year={2018},
publisher={MIT Press}
}

@article{gaiSurvey,
title={Generative Artificial Intelligence in Robotic Manipulation: A Survey},
author={Kun Zhang and Peng Yun and Jun Cen and others},
journal={CoRR},
volume={abs/2503.03464},
year={2025}
}

@article{vinyals2019grandmaster,
title={Grandmaster level in StarCraft II using multi-agent reinforcement learning},
author={Vinyals, Oriol and Babuschkin, Igor and Czarnecki, Wojciech M and Mathieu, Micha{\"e}l and Dudzik, Andrew and Chung, Junyoung and Choi, David H and Powell, Richard and Ewalds, Timo and Georgiev, Petko and others},
journal={Nature},
volume={575},
number={7782},
pages={350--354},
year={2019},
publisher={Nature Publishing Group}
}

@article{mnih2015human,
title={Human-level control through deep reinforcement learning},
author={Mnih, Volodymyr and Kavukcuoglu, Koray and Silver, David and Rusu, Andrei A and Veness, Joel and Bellemare, Marc G and Graves, Alex and Riedmiller, Martin and Fidjeland, Andreas K and Ostrovski, Georg and others},
journal={Nature},
volume={518},
number={7540},
pages={529--533},
year={2015},
publisher={Nature Publishing Group}
}

@inproceedings{kaelbling2011hierarchical,
title={Hierarchical task and motion planning in the now},
author={Kaelbling, Leslie Pack and Lozano-P{\'e}rez, Tom{\'a}s},
booktitle={IEEE International Conference on Robotics and Automation},
pages={1470--1477},
year={2011},
organization={IEEE}
}

@InProceedings{pmlr-v202-driess23a,
title={{P}a{LM}-E: An Embodied Multimodal Language Model},
author={Driess, Danny and Xia, Fei and Sajjadi, Mehdi S. M. and others},
booktitle={Proceedings of the International Conference on Machine Learning},
pages={8469--8488},
year={2023},
editor={Krause, Andreas et al.},
volume={202},
series={Proceedings of Machine Learning Research},
month={23--29 Jul},
publisher={PMLR}
}

@article{tsesmelis2024re,
title={Re-assembling the past: The RePAIR dataset and benchmark for real world 2D and 3D puzzle solving},
author={Tsesmelis, Theodore and Palmieri, Luca and Khoroshiltseva, Marina and others},
journal={Advances in Neural Information Processing Systems},
volume={37},
pages={30076--30105},
year={2024}
}

@inproceedings{gao2022fast,
title={Fast high-quality tabletop rearrangement in bounded workspace},
author={Gao, Kai and Lau, Darren and Huang, Baichuan and Bekris, Kostas E and Yu, Jingjin},
booktitle={IEEE International Conference on Robotics and Automation},
pages={1961--1967},
year={2022},
organization={IEEE}
}

@inproceedings{baum2017opening,
title={Opening a lockbox through physical exploration},
author={Baum, Manuel and Bernstein, Matthew and Martin-Martin, Roberto and others},
booktitle={IEEE International Conference on Humanoid Robots},
pages={461--467},
year={2017},
organization={IEEE}
}

@article{kingston2019exploring,
title={Exploring implicit spaces for constrained sampling-based planning},
author={Kingston, Zachary and Moll, Mark and Kavraki, Lydia E},
journal={International Journal of Robotics Research},
volume={38},
number={10-11},
pages={1151--1178},
year={2019},
publisher={SAGE Publications Sage UK: London, England}
}

@article{garrett2021integrated,
title={Integrated task and motion planning},
author={Garrett, Caelan Reed and Chitnis, Rohan and Holladay, Rachel and others},
journal={Annual Review of Control, Robotics, and Autonomous Systems},
volume={4},
number={1},
pages={265--293},
year={2021},
publisher={Annual Reviews}
}

@article{bellemare2013arcade,
title={The arcade learning environment: An evaluation platform for general agents},
author={Bellemare, Marc G and Naddaf, Yavar and Veness, Joel and Bowling, Michael},
journal={Journal of Artificial Intelligence Research},
volume={47},
pages={253--279},
year={2013}
}

@inproceedings{samvelyan2021minihack,
title={MiniHack the Planet: A Sandbox for Open-Ended Reinforcement Learning Research},
author={Mikayel Samvelyan and Robert Kirk and Vitaly Kurin and others},
booktitle={ Neural Information Processing Systems},
year={2021},
}

@inproceedings{toussaint2018differentiable,
title={Differentiable physics and stable modes for tool-use and manipulation planning},
author={Toussaint, Marc A and Allen, Kelsey Rebecca and Smith, Kevin A and Tenenbaum, Joshua B},
year={2018},
booktitle={Proceedings of Robotics: Science and Systems}
}

@article{firoozi2025foundation,
title={Foundation models in robotics: Applications, challenges, and the future},
author={Firoozi, Roya and Tucker, Johnathan and Tian, Stephen and others},
journal={International Journal of Robotics Research},
volume={44},
number={5},
pages={701--739},
year={2025},
publisher={SAGE Publications Sage UK: London, England}
}

@article{lee2018dart,
title={Dart: Dynamic Animation and Robotics Toolkit},
author={Lee, Jeongseok and Grey, Michael X and Ha, Sehoon and others},
journal={Journal of Open Source Software},
volume={3},
number={22},
pages={500},
year={2018},
publisher={The Open Journal}
}

@book{conlan2017blender,
title={The Blender Python API: Precision 3D Modeling and Add-on Development},
author={Conlan, Chris},
year={2017},
publisher={Apress}
}

@inproceedings{kuffner2000rrt,
title={RRT-Connect: An Efficient Approach to Single-Query Path Planning},
author={Kuffner, James J and LaValle, Steven M},
booktitle={IEEE International Conference on Robotics and Automation},
volume={2},
pages={995--1001},
year={2000},
organization={IEEE}
}

@inproceedings{kartal2016data,
title={Data-Driven Sokoban Puzzle Generation with Monte Carlo Tree Search},
author={Kartal, Bilal and Sohre, Nick and Guy, Stephen J},
booktitle={Proceedings of the AAAI Conference on Artificial Intelligence and Interactive Digital Entertainment},
volume={12},
pages={58--64},
year={2016}
}

@article{simeon2004manipulation,
title={Manipulation Planning with Probabilistic Roadmaps},
author={Sim{\'e}on, Thierry and Laumond, Jean-Paul and Cort{\'e}s, Juan and Sahbani, Anis},
journal={International Journal of Robotics Research},
volume={23},
number={7--8},
pages={729--746},
year={2004},
publisher={SAGE Publications}
}

@book{taylor2015procedural,
title={The Procedural Generation of Interesting Sokoban Levels},
author={Taylor, Joshua},
year={2015},
publisher={University of North Texas}
}

@book{harris2020exploring,
title={Exploring Roguelike Games},
author={Harris, John},
year={2020},
publisher={CRC Press}
}

@mastersthesis{karman2018generating,
title={Generating Sokoban Levels That Are Interesting to Play Using Simulation},
author={Karman, SJ and others},
year={2018},
school={Utrecht University}
}

@article{awiszus2022wor,
title={Wor(l)d-GAN: Toward Natural-Language-Based PCG in Minecraft},
author={Awiszus, Maren and Schubert, Frederik and Rosenhahn, Bodo},
journal={IEEE Transactions on Games},
volume={15},
number={2},
pages={182--192},
year={2022},
publisher={IEEE}
}

@inproceedings{herve2021comparing,
title={Comparing PCG Metrics with Human Evaluation in Minecraft Settlement Generation},
author={Herv{\'e}, Jean-Baptiste and Salge, Christoph},
booktitle={Proceedings of the International Conference on the Foundations of Digital Games},
pages={1--15},
year={2021}
}

@article{zakaria2022procedural,
title={Procedural Level Generation for Sokoban via Deep Learning: An Experimental Study},
author={Zakaria, Yahia and Fayek, Magda and Hadhoud, Mayada},
journal={IEEE Transactions on Games},
volume={15},
number={1},
pages={108--120},
year={2022},
publisher={IEEE}
}

@inproceedings{kingston2022robowflex,
title={Robowflex: Robot Motion Planning with Moveit Made Easy},
author={Kingston, Zachary and Kavraki, Lydia E},
booktitle={IEEE/RSJ International Conference on Intelligent Robots and Systems},
pages={3108--3114},
year={2022},
organization={IEEE}
}

@inproceedings{jia2024cluttergen,
title={Cluttergen: A Cluttered Scene Generator for Robot Learning},
author={Jia, Yinsen and Chen, Boyuan},
booktitle={Proceedings of the Conference on Robot Learning},
year={2024}
}

@article{kober2013reinforcement,
title={Reinforcement Learning in Robotics: A Survey},
author={Kober, Jens and Bagnell, J Andrew and Peters, Jan},
journal={International Journal of Robotics Research},
volume={32},
number={11},
pages={1238--1274},
year={2013},
publisher={SAGE Publications}
}

@article{dekegel2019procedural,
title={Procedural Puzzle Generation: A Survey},
author={De Kegel, Barbara and Haahr, Mads},
journal={IEEE Transactions on Games},
volume={12},
number={1},
pages={21--40},
year={2019},
publisher={IEEE}
}

@inproceedings{togelius2011procedural,
title={What Is Procedural Content Generation? Mario on the Borderline},
author={Togelius, Julian and Kastbjerg, Emil and Schedl, David and Yannakakis, Georgios N},
booktitle={International Workshop on Procedural Content Generation in Games},
pages={1--6},
year={2011}
}

@article{xiao2025robot,
title={Robot Learning in the Era of Foundation Models: A Survey},
author={Xiao, Xuan and Liu, Jiahang and Wang, Zhipeng and others},
journal={Neurocomputing},
pages={129963},
year={2025},
publisher={Elsevier}
}

@book{craig2017introduction,
title={Introduction to Robotics: Mechanics and Control},
author={Craig, John J},
edition={4},
year={2017},
publisher={Pearson},
address={Upper Saddle River, NJ},
isbn={978-0133489798}
}

@article{hendrikx2013procedural,
title={Procedural Content Generation for Games: A Survey},
author={Hendrikx, Mark and Meijer, Sebastiaan and Van Der Velden, Joeri and Iosup, Alexandru},
journal={ACM Transactions on Multimedia Computing, Communications, and Applications},
volume={9},
number={1},
pages={1--22},
year={2013},
publisher={ACM}
}

@book{pcgbook,
title={Procedural Content Generation in Games: A Textbook and an Overview of Current Research},
author={Shaker, Noor and Togelius, Julian and Nelson, Mark J},
publisher={Springer},
year={2016}
}

@inproceedings{cobbe2020leveraging,
title={Leveraging Procedural Generation to Benchmark Reinforcement Learning},
author={Cobbe, Karl and Hesse, Chris and Hilton, Jacob and Schulman, John},
booktitle={Proceedings of the International Conference on Machine Learning},
pages={2048--2056},
year={2020},
organization={PMLR}
}

@article{viana2021procedural,
title={Procedural Dungeon Generation: A Survey},
author={Viana, Breno M F and dos Santos, Selan R},
journal={Journal on Interactive Systems},
volume={12},
number={1},
pages={83--101},
year={2021}
}

@article{culberson1997sokoban,
title={Sokoban Is PSPACE-Complete},
author={Culberson, Joseph},
journal={IEICE Technical Report},
year={1997}
}

@article{openai2019rubiks,
title={Solving Rubik's Cube with a Robot Hand},
author={OpenAI and Akkaya, Ilge and Andrychowicz, Marcin and others},
journal={ArXiv Preprint},
year={2019}
}

@article{deitke2022procthor,
title={ProcTHOR: Large-Scale Embodied AI Using Procedural Generation},
author={Deitke, Matt and VanderBilt, Eli and Herrasti, Alvaro and others},
journal={Advances in Neural Information Processing Systems},
volume={35},
pages={5982--5994},
year={2022}
}

@inproceedings{yu2020meta,
title={Meta-World: A Benchmark and Evaluation for Multi-Task and Meta Reinforcement Learning},
author={Yu, Tianhe and Quillen, Deirdre and He, Zhanpeng and others},
booktitle={Proceedings of the Conference on Robot Learning},
pages={1094--1100},
year={2020},
organization={PMLR}
}

@article{james2020rlbench,
title={RLBench: The Robot Learning Benchmark and Learning Environment},
author={James, Stephen and Ma, Zicong and Arrojo, David Rovick and Davison, Andrew J},
journal={IEEE Robotics and Automation Letters},
volume={5},
number={2},
pages={3019--3026},
year={2020},
publisher={IEEE}
}

@article{moll2015benchmarking,
title={Benchmarking Motion Planning Algorithms: An Extensible Infrastructure for Analysis and Visualization},
author={Moll, Mark and Sucan, Ioan A and Kavraki, Lydia E},
journal={IEEE Robotics and Automation Magazine},
volume={22},
number={3},
pages={96--102},
year={2015},
publisher={IEEE}
}

@article{sucan2012open,
title={The Open Motion Planning Library},
author={Sucan, Ioan A and Moll, Mark and Kavraki, Lydia E},
journal={IEEE Robotics and Automation Magazine},
volume={19},
number={4},
pages={72--82},
year={2012},
publisher={IEEE}
}

@misc{coumans2021pybullet,
title={PyBullet, a Python Module for Physics Simulation for Games, Robotics and Machine Learning},
author={Coumans, Erwin and Bai, Yunfei},
year={2021}
}

@misc{blender,
title={Blender},
author={Blender Online Community},
organization={Blender Foundation},
address={Stichting Blender Foundation, Amsterdam},
year={2021}
}

@article{phobos2020,
title={Phobos: A Tool for Creating Complex Robot Models},
author={von Szadkowski, Kai and Reichel, Simon},
journal={Journal of Open Source Software},
volume={5},
number={45},
pages={1326},
year={2020}
}

@article{Orthey2023AnnualReview,
title={Sampling-Based Motion Planning: A Comparative Review},
author={Orthey, Andreas and Chamzas, Constantinos and Kavraki, Lydia E},
journal={Annual Review of Control, Robotics, and Autonomous Systems},
volume={7},
number={1},
pages={285--310},
year={2024}
}

@article{Bayraktar2023RAL,
title={Solving Rearrangement Puzzles Using Path Defragmentation in Factored State Spaces},
author={Bayraktar, Servet B and Orthey, Andreas and Kingston, Zachary and Toussaint, Marc and Kavraki, Lydia E},
journal={IEEE Robotics and Automation Letters},
volume={8},
number={8},
pages={4529--4536},
year={2023}
}

@article{gammell2020batch,
title={Batch Informed Trees (BIT*): Informed Asymptotically Optimal Anytime Search},
author={Gammell, Jonathan D and Barfoot, Timothy D and Srinivasa, Siddhartha S},
journal={International Journal of Robotics Research},
volume={39},
number={5},
pages={543--567},
year={2020},
publisher={SAGE Publications}
}

@article{salzman2016asymptotically,
title={Asymptotically Near-Optimal RRT for Fast, High-Quality Motion Planning},
author={Salzman, Oren and Halperin, Dan},
journal={IEEE Transactions on Robotics},
volume={32},
number={3},
pages={473--483},
year={2016},
publisher={IEEE}
}

@article{gammell2021asymptotically,
title={Asymptotically Optimal Sampling-Based Motion Planning Methods},
author={Gammell, Jonathan D and Strub, Marlin P},
journal={Annual Review of Control, Robotics, and Autonomous Systems},
volume={4},
number={1},
pages={295--318},
year={2021},
publisher={Annual Reviews}
}

@article{karaman2011sampling,
title={Sampling-Based Algorithms for Optimal Motion Planning},
author={Karaman, Sertac and Frazzoli, Emilio},
journal={International Journal of Robotics Research},
volume={30},
number={7},
pages={846--894},
year={2011},
publisher={SAGE Publications}
}

@inproceedings{Kamat2022IROS,
title={BITKOMO: Combining Sampling and Optimization for Fast Convergence in Optimal Motion Planning},
author={Kamat, Jay and Ortiz-Haro, Joaquim and Toussaint, Marc and Pokorny, Florian T and Orthey, Andreas},
booktitle={International Conference on Intelligent Robots and Systems},
pages={4492--4497},
year={2022}
}

@article{Orthey2021TRO,
title={Section Patterns: Efficiently Solving Narrow Passage Problems in Multilevel Motion Planning},
author={Orthey, Andreas and Toussaint, Marc},
journal={IEEE Transactions on Robotics},
volume={37},
pages={1891--1905},
year={2021}
}

@article{Chamzas2022RAL,
	title = {MotionBenchMaker: A Tool to Generate and Benchmark Motion Planning Datasets},
	volume = {7},
	number = {2},
	pages = {882--889},
	issn = {2377-3766},
	journal = {IEEE Robotics and Automation Letters},
	author = {Chamzas, Constantinos and Quintero-Pe{\~n}a, Carlos and Kingston, Zachary and Orthey, Andreas and Rakita, Daniel and Gleicher, Michael and Toussaint, Marc and E. Kavraki, Lydia},
	year = {2022},
	month = {apr},
	month_numeric = {4},
}
}
\end{document}